\documentclass[letterpaper]{article}
\usepackage{arxivtemplate}
\usepackage{times}
\usepackage{helvet}
\usepackage{courier}
\usepackage[hyphens]{url}
\usepackage{graphicx}
\usepackage{natbib}
\usepackage{caption}
\usepackage{algorithm}
\usepackage{algorithmic}
\usepackage{newfloat}
\usepackage{listings}
\DeclareCaptionStyle{ruled}{labelfont=normalfont,labelsep=colon,strut=off}
\floatstyle{ruled}
\newfloat{listing}{tb}{lst}{}
\floatname{listing}{Listing}
\newcommand{\customsmall}{\fontsize{8.3}{5.5}\selectfont}
\newcommand{\customsmallB}{\fontsize{8.3}{7.5}\selectfont}

\usepackage{amsmath,amsfonts,bm}

\def\eqref#1{equation~\ref{#1}}

\def\1{\bm{1}}

\def\vh{{\bm{h}}}

\def\vv{{\bm{v}}}

\def\vx{{\bm{x}}}

\def\mG{{\bm{G}}}

\def\mX{{\bm{X}}}

\DeclareMathAlphabet{\mathsfit}{\encodingdefault}{\sfdefault}{m}{sl}
\SetMathAlphabet{\mathsfit}{bold}{\encodingdefault}{\sfdefault}{bx}{n}

\def\sN{{\mathbb{N}}}

\def\sR{{\mathbb{R}}}

\def\sV{{\mathbb{V}}}

\usepackage{booktabs}
\usepackage{makecell}
\usepackage{cleveref}

\title{Functional Connectivity Graph Neural Networks}
\author {
    Yang Li\textsuperscript{\rm 1},
    Luopeiwen Yi\textsuperscript{\rm 2},
    Tananun Songdechakraiwut\textsuperscript{\rm 1}
}
\affiliations {
    \textsuperscript{\rm 1}Department of Computer Science, Duke University, Durham, NC, USA\\
    \textsuperscript{\rm 2}Social Science Research Institute, Duke University, Durham, NC, USA
}

\begin{document}

\maketitle

\begin{abstract}
Real-world networks often benefit from capturing both local and global interactions. Inspired by multi-modal analysis in brain imaging, where structural and functional connectivity offer complementary views of network organization, we propose a graph neural network framework that generalizes this approach to other domains. Our method introduces a functional connectivity block based on persistent graph homology to capture global topological features. Combined with structural information, this forms a multi-modal architecture called Functional Connectivity Graph Neural Networks. Experiments show consistent performance gains over existing methods, demonstrating the value of brain-inspired representations for graph-level classification across diverse networks.
\end{abstract}

\section{Introduction}

\begin{figure*}[t]
\centering
\includegraphics[width=\textwidth]{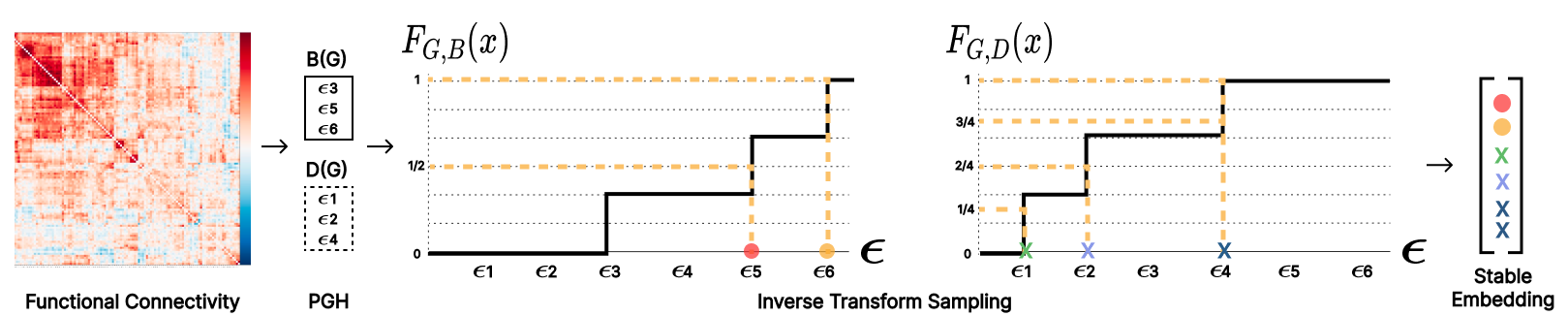}
\caption{A schematic of the functional connectivity block. First, functional connectivity is extracted from a structural graph. Persistent graph homology (PGH) is then used to extract topological invariants, which are then represented as the birth and death sets. Inverse transform sampling is finally used to upsample or downsample the PGH representation into a consistent dimensionality across graphs of varying sizes, resulting in a stable embedding.}
\label{fig:fcBlock}
\end{figure*}

Real-world networks, spanning social, biological, and technological systems, exhibit topologies distinct from purely regular or random configurations \cite{kossinets2006empirical, bullmore2012economy, farahani2013review}. The brain exemplifies this complexity and robustness, integrating structural and functional connectivity as revealed by advanced imaging techniques such as fMRI and dMRI \cite{mulkern2006complementary, leergaard2012mapping, colclough2017heritability, babaeeghazvini2021brain}. Together, these modalities provide a comprehensive picture of how the brain balances localized specialization with global integration for efficient information processing and adaptability \cite{bassett_small-world_2006, lv2023synergy}. This synergy and resilience serve as inspiration for our proposed graph learning model, which, while informed by neuroscience, is designed for application to general network data beyond the brain.

This integrative perspective motivates the use of graph-based representations for analyzing complex systems, as they provide a flexible framework for encoding and systematically exploring relationships within a wide range of networks \cite{zhang2018network, jiang2021could}. In particular, multi-modal integration, which combines complementary forms of connectivity, can capture topological nuances more effectively than single-modality approaches \cite{songdechakraiwut2021topological}. Graph neural networks (GNNs) have emerged as powerful tools for processing such data, propagating information between adjacent nodes according to the underlying structural edges \cite{scarselli2008graph, xu2018powerful, wu2020comprehensive, wu2022graph}. However, traditional GNNs often rely solely on structural connectivity, limiting their ability to capture the intrinsic functional architecture that shapes many real-world networks. Functional connections represent long-range interactions through co-activation or co-variation, providing a perspective distinct from, but complementary to, structural connections and highlighting the importance of functional synergy in network organization.

Building on this foundation, we broaden the scope of graph analysis by explicitly integrating neuroimaging-inspired functional connectivity into the study of diverse real-world networks. To this end, we introduce a modular functional connectivity block for neural network architectures. This block processes structural graphs as input and extracts their functional topology using persistent graph homology \cite{songdechakraiwut2023topological}, yielding mathematically stable representations of topological invariants. Based on this approach, we propose functional connectivity graph neural networks (FC-GNNs), a unified architecture that integrates both functional and structural modalities. Evaluation across 13 biological and social network datasets \cite{morris2020tudataset} demonstrates significant performance improvements, highlighting the cross-domain value of incorporating functional topology.

This work translates brain connectivity principles into universal graph learning. Unlike domain-specific neuro-studies \cite{kim2020understanding, songdechakraiwut2020dynamic, li2021braingnn, mahmood2021deep, lei2022graph, pitsik2023topology, gu2024novel, wang2024ifc}, our work leverages decades of insights from neuroscience and connectomics, adapting these principles to apply broadly across all types of graph data using the proposed functional connectivity block. The resulting FC-GNNs effectively unify functional and structural connectivity, advancing graph-level classification tasks. This new multi-modal approach bridges the gap between brain-specific research and broader graph representation learning, providing a unified methodology for analyzing a broad category of complex networks.

\begin{figure*}[t]
\centering
\includegraphics[width=\textwidth]{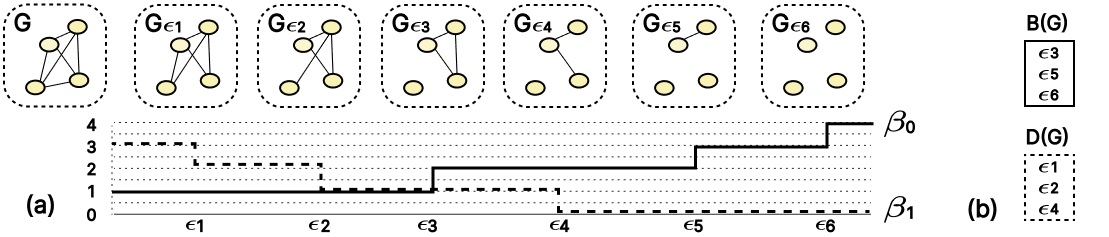}
\caption{(a) Graph filtration of a four-node network \( G \). As the filtration value \( \epsilon \) increases, the number of connected components ($\beta_0$) increases while the number of cycles ($\beta_1$) decreases, one at a time. New connected components are born at edge weights \( \epsilon_3, \epsilon_5, \epsilon_6 \), while existing cycles die at \( \epsilon_1, \epsilon_2, \epsilon_4 \). Thus, $B(G) = \{\epsilon_3, \epsilon_5, \epsilon_6\}$ and $D(G) = \{\epsilon_1, \epsilon_2, \epsilon_4\}$, as shown in (b).
}
\label{fig:betti}
\end{figure*}

\section{Functional Connectivity Block} \label{sec:fcb}

The functional connectivity block comprises two modules: the connectivity feature module, which captures functional relationships; and the topological feature module, which identifies higher-order attributes of these relationships. This block is versatile and can be positioned flexibly within a neural network architecture to process structural graphs as input and produce high-level abstractions of functional topology. Figure \ref{fig:fcBlock} presents a schematic illustration of the functional connectivity block.

\subsection{Connectivity feature module}

Let $\mG_{struct} = (V, E)$ be a structural graph, where $V$ is the set of nodes representing distinct entities, and $E$ is the set of edges representing connections between these entities. Each edge $E$ in $\mG_{struct}$ denotes a direct relationship or interaction between two nodes in $V$, which may represent structural links or other forms of connectivity depending on the context of the graph. Real-world structural graphs represent networks derived from various phenomena. For example, in structural brain networks derived from diffusion tensor imaging (DTI), nodes represent distinct regions of the human brain, and edges represent physical connections in the form of white matter tracts \cite{lazar2010mapping, lim2015preferential}. These physical connections reflect the anatomical pathways through which neurons transmit signals, enabling communication and coordination across different areas of the brain \cite{lazar2010mapping}.

Without loss of generality, we assume that each node $i \in V$ has a node feature vector $\vx_i = (x_{i1},x_{i2},...,x_{id}) \in \sR^d$ associated with it. Note that these attributes can be intrinsically available, derived from node-level graph theoretical summaries (such as degree or centrality measures) \cite{borgatti2006graph}, or learned through graph neural networks (GNNs) by leveraging the structure of the graph \cite{wu2019net, wu2020comprehensive}.
When two node features $\vx_i, \vx_j$ from a pair of nodes $i, j$ are statistically dependent, they exhibit functional synergy. This synergy can be quantified using various measures of statistical dependency, including Pearson correlation, partial correlation, Spearman's rank correlation, and mutual information \cite{baba2004partial, zar2005spearman, cohen2009pearson, epskamp2018tutorial}. In macroscale connectomics, Pearson correlation is widely used to compute functional connectivity of the human brain because it effectively detects synchrony between brain regions, a hallmark of functional neural activity \cite{liang2012effects, zalesky2012connectivity, abdelnour2014network, jin2017dynamic, zhan2017significance}. Its interpretation is straightforward: values range from -1 to 1, with values farther from 0 indicating stronger functional connections. Its computation is efficient, making it well-suited for large-scale graph datasets \cite{cohen2009pearson, liang2012effects}. Therefore, Pearson correlation is used to construct functional connectivity in this work due to its interpretability and computational efficiency.

Let $\mX = (x_{ij})_{1\leq i \leq |V|,\;1 \leq j \leq d}$ be a node feature matrix, where  $|V|$ denotes the number of nodes in $\mG_{struct}$, and $d$ denotes the dimensionality of the node feature vectors. Formally, Pearson correlation between nodes $i$ and $j$ is defined as
$r_{ij} = \mathrm{conv}(\vx_i - \bar{\vx}_i,\, \vx_j - \bar{\vx}_j) / \big( \sigma(\vx_i)\,\sigma(\vx_j) \big)$, where $\vx_i$ and $\vx_j$ denote the $i$-th and $j$-th rows of $\mX$, $\bar{\vx}_i$ is the mean of $\vx_i$, $\mathrm{conv}$ denotes the dot product, and $\sigma(\vx_i)$ denotes the standard deviation of $\vx_i$.
We define a $|V|$-by-$|V|$ weighted adjacency matrix $\mG_{fc} = (r_{ij})$ that captures the pairwise Pearson correlations for \emph{all} pairs of nodes, encoding interactions that extend beyond basic spatial or structural adjacency  \cite{coscia2021pearson}. This interaction is commonly referred to as functional connectivity, representing the relationship between spatially separated nodes that share functional properties. For instance, in the human brain, structural connectivity typically identifies physical connections between adjacent regions \cite{westin2007extracting, ghosh2015survey, grier2020estimating}, while functional connectivity highlights similar patterns of activation across different regions, irrespective of physical connectedness \cite{rogers2007assessing, van2010exploring, bijsterbosch2017introduction}.
We refer to this matrix $\mG_{fc}$ as \emph{functional connectivity}. Note that the Pearson correlation for a node with itself is always 1, and is typically excluded from analyses, resulting in $\mG_{fc}$ being a weighted graph with no self-loops.

\subsection{Topological feature module} \label{subsec:top_feature_module}

The functional connectivity $\mG_{fc}$ is a dense matrix, often representing a \emph{complete} graph. Consequently, standard workflows \cite{bullmore2009complex} apply thresholds to correlation values for two primary purposes: to produce a sparser graph with a clearer structure and to reduce the computational runtime of analytical methods. Persistent homology methods, for example, exhibit cubic time complexity \cite{otter2017roadmap}, rendering them infeasible for large, complete graphs without approximation. These approximations can introduce numerical errors and degrade the signal-to-noise ratio. Moreover, varying threshold values significantly alter the graph's structure, potentially influencing the study's conclusions. Fixed spatial thresholds, in particular, limit the range of functional connectivity that can be analyzed.

To address these challenges, we utilize \emph{persistent graph homology}, a scalable learning framework that enables topological analyses of large-scale functional connectivity through closed-form computation \cite{songdechakraiwut2023wasserstein}. This approach has emerged as an effective tool for understanding, characterizing, and quantifying human connectomes \cite{songdechakraiwut2022fast}. Persistent graph homology captures interpretable topological invariants, including connected components (0D topological features) and cycles (1D topological features), across the full threshold resolution of a graph.

Formally, given a functional connectivity $\mG_{fc}$, we derive a binary graph $\mG_{fc,\epsilon}$ by applying a threshold $\epsilon$ to the edge correlations, so that an edge between nodes $i$ and $j$ exists if $r_{ij} > \epsilon$. As $\epsilon$ increases, edges are progressively removed from  $\mG_{fc}$. This process results in a filtration \cite{lee2012persistent}:
$\mG_{fc,\epsilon_0} \supseteq \mG_{fc,\epsilon_1} \supseteq \cdots \supseteq \mG_{fc,\epsilon_k},$
where $\epsilon_0 \leq \epsilon_1 \leq \cdots \leq \epsilon_k$ are called filtration values.
Persistent homology describes how topological features are born and die as filtration values  $\epsilon$ change. A feature that is born at filtration $b_l$ and dies at filtration $d_l$ is represented by a point $(b_l, d_l)$ on a two-dimensional plane. The collection of all such points ${(b_l, d_l)}$ forms what is known as a \emph{persistence diagram} \cite{edelsbrunner2022computational}. As $\epsilon$ increases, the number of connected components $\beta_0(\mG_{fc,\epsilon})$ (0th Betti) increases monotonically, while the number of cycles $\beta_1(\mG_{fc,\epsilon})$ (1st Betti) decreases monotonically. Therefore, persistent graph homology only needs to track the birth values $B(\mG_{fc})$ for connected components and the death values $D(\mG_{fc})$ for cycles, given as \cite{songdechakraiwut2023topological}
\begin{align} \label{eq:bd_decomp}
    B(\mG_{fc}) = \{b_l \}_{l=1}^{|V|-1}, \quad D(\mG_{fc})=\{ d_l \}_{l=1}^{1 + |V| (|V| - 3)/2}.
\end{align}
Figure \ref{fig:betti} illustrates an example of the graph filtration on a four-node network, and its corresponding birth and death sets.

Underlyingly, an empirical distribution for the persistence diagram of connected components in $\mG_{fc}$ is defined using Dirac masses as \cite{turner2014frechet}
\begin{equation*}
    f_{\mG_{fc},B}(x) := \frac{1}{|B(\mG_{fc})|} \sum_{b \in B(\mG_{fc})} \delta(x-b),
\end{equation*}
where $\delta(x-b)$ is a Dirac delta centered at the point $b$. The empirical function is then defined as the integral of $f_{\mG_{fc},B}$ expressed as
\begin{equation*}
    F_{\mG_{fc},B}(x) = \frac{1}{|B(\mG_{fc})|} \sum_{b \in B(\mG_{fc})} \1_{b \leq x},
\end{equation*}
where $\1_{b \leq x}$ is an indicator function that takes the value 1 if $b \leq x$, and 0 otherwise.
A pseudo-inverse of $F_{\mG_{fc},B}$ is defined as
\begin{equation} \label{eq:inversionsampling}
    F_{\mG_{fc},B}^{-1}(z) = \inf \{b \in \mathbb{R}\,|\, F_{\mG_{fc},B}(b) \geq z\},
\end{equation}
i.e., $F_{\mG_{fc},B}^{-1}(z)$ is the smallest value of $b$ such that $F_{\mG_{fc},B}(b) \geq z$.

\begin{figure*}[!t]
\centering
\includegraphics[width=\textwidth]{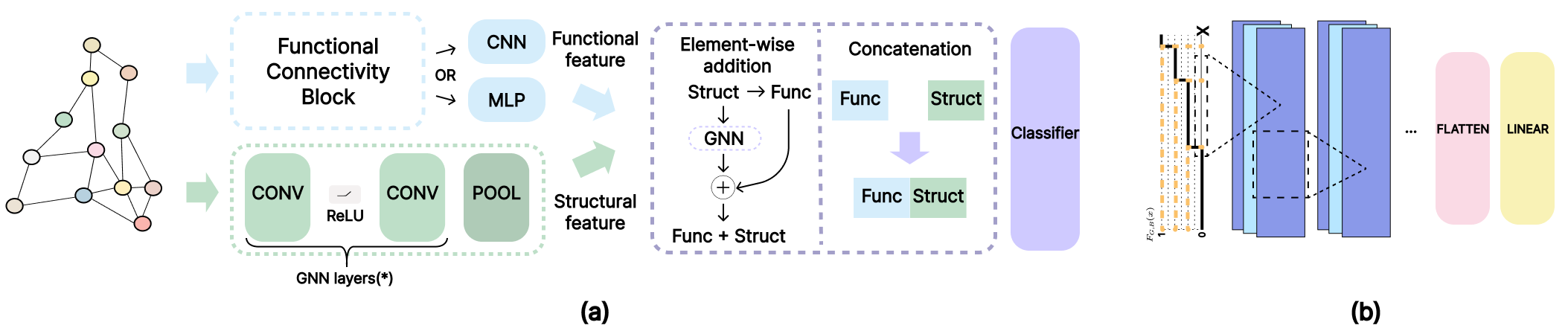}
\caption{(a) A schematic of functional connectivity graph neural networks. The input graph is processed through two parallel pipelines. The top pipeline extracts functional connectivity, as detailed in Section \ref{sec:fcb}, and performs representation learning using the proposed MLP or 1D CNN to extract functional features. The bottom pipeline extracts structural features through a GNN, which consists of graph convolution layers, ReLU activations, and pooling layers. The notation (*) indicates that each block may contain one or more combinations of layers to allow for adaptable model complexity. These functional and structural features are then integrated via element-wise addition or concatenation for classification tasks. (b) An illustration of how 1D convolutional layers process stair shapes in the Betti curve, which encodes the spatial relationships between adjacent values within topological embeddings.}
\label{fig:pipeline}
\end{figure*}

Consider $\mG_{fc}^{(1)},\mG_{fc}^{(2)},...,\mG_{fc}^{(N)}$, a set of $N$ observed functional connectivities potentially with varying node sizes, i.e., their birth and death sets may differ in size, as detailed in Eq. (\ref{eq:bd_decomp}). Let $F^{-1}_{\mG_{fc}^{(i)},B}$ be the pseudo-inverse of network $\mG_{fc}^{(i)}$.
To embed the persistence diagram of connected components from network $\mG_{fc}^{(i)}$, we define a vector by sampling the pseudo-inverse at regular intervals $\frac{1}{m},\frac{2}{m},...,\frac{m}{m}$:
\begin{align*}
    \vv_{B,i} := 
    \big( F^{-1}_{\mG_{fc}^{(i)},B}(\frac{1}{m}),F^{-1}_{\mG_{fc}^{(i)},B}(\frac{2}{m}),...,F^{-1}_{\mG_{fc}^{(i)},B}(\frac{m}{m}) \big)^\top .
\end{align*}
Similarly, the embedding of a persistence diagram for cycles in network $\mG_{fc}^{(i)}$ is defined as a vector obtained by sampling the pseudo-inverse at regular intervals $\frac{1}{n},\frac{2}{n},...,\frac{n}{n}$:
\begin{align*}
    \vv_{D,i} := 
    \big( F^{-1}_{\mG_{fc}^{(i)},D}(\frac{1}{n}), F^{-1}_{\mG_{fc}^{(i)},D}(\frac{2}{n}), ..., F^{-1}_{\mG_{fc}^{(i)},D}(\frac{n}{n})\big)^\top .
\end{align*}

\paragraph{Scalability.} Computing $\vv_{B,i}$ and $\vv_{D,i}$ mainly involves finding the birth set $B(\mG_{fc})$ (edges in the maximum spanning tree, computed in $O(n \log n)$ time) and the death set $D(\mG_{fc})$ (the remaining edges). Thus, both vectors can be efficiently obtained in $O(n \log n)$ time, where $n$ is the number of edges in the functional connectivity graph.

\begin{figure*}[t]
\centering
\includegraphics[width=\textwidth]{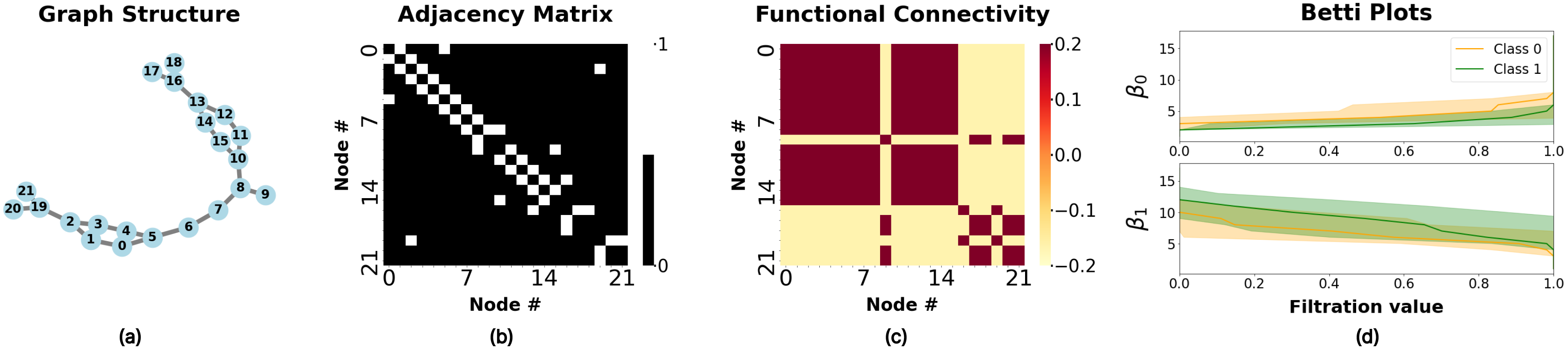}
\caption{(a), (b), and (c): Visualization of a sample representative from MUTAG (Class 0). The graph structure is depicted with nodes and edges, where nodes are labeled by their indices. The graph structure is a binary, undirected graph represented using an adjacency matrix, where entries of 0 indicate the absence of an edge, and entries of 1 indicate the presence of an edge between nodes. The functional connectivity of this sample is computed using the pairwise Pearson correlation between node features. (d): Betti plots based on all samples in MUTAG, with the top plot representing the 0th Betti number (number of connected components) and the bottom plot representing the 1st Betti number (number of cycles). Thick lines indicate the Wasserstein barycenters, while the shaded areas around the barycenters represent the standard deviation, differentiating between Class 0 and Class 1, the class labels of MUTAG.}
\label{fig:schematic}
\end{figure*}

\section{Functional Connectivity GNNs}

Here, we propose new neural networks for complex graphs that learn both structural and functional information through the integration of a functional connectivity block and a graph neural network block, as detailed below. We call this architecture \emph{functional connectivity graph neural networks}. Figure \ref{fig:pipeline} presents its schematic illustration, which is detailed in the following sections.

\subsection{Functional representation learning} \label{sec:mandn}

The topological embeddings $\vv_{B,i}$ and $\vv_{D,i}$, defined in Section \ref{subsec:top_feature_module}, are stable and enable functional connectivity of graphs of varying sizes to be embedded into a consistent dimensionality, as determined by the hyperparameters $m$ and $n$. These embeddings are readily usable for downstream representation learning.

The hyperparameters $m$ and $n$ can be chosen using dataset metadata or through data-driven optimization approaches. One metadata-based option is to set $m$ and $n$ based on the cardinality of $B(\mG_{fc})$ and $D(\mG_{fc})$ from the graph with the largest number of nodes in the dataset. This approach involves upsampling the birth and death sets of smaller graphs to match the largest graph, ensuring no loss of information and leveraging the entire dataset. Alternatively, $m$ and $n$ can be determined using the smallest graph in the dataset, downsampling larger graphs to match it. While this reduces the data needed for training generalized models, it may result in some information loss.

A balanced approach is to set $m$ and $n$ to the average number of nodes and edges across the dataset. This middle ground preserves a reasonable amount of information while maintaining computational efficiency. Ultimately, these hyperparameters can also be optimized through data-driven methods, such as grid search or random search, to identify the optimal values for $m$ and $n$ for a given task.

To learn from these topological embeddings, we propose two approaches, which are detailed below. 

\paragraph{Approach 1.} The first approach leverages a multi-layer perceptron (MLP), a feed-forward neural network architecture that is appropriate for small to moderate-sized graphs \cite{ruck1990feature}. The rationale behind using MLP is its ability to process fixed-size embeddings, capture high-level abstractions, and deliver reliable performance for tasks like classification \cite{ramchoun2016multilayer, almeida2020multilayer}. Conceptually, the MLP processes the concatenated topological embeddings to produce a high-level feature vector:
$\vh_{\text{MLP}} = \text{MLP}([\vv_{B,i}; \vv_{D,i}])$,
where MLP$(\cdot)$ comprises fully connected layers, bias terms, and nonlinear activation functions, and $[\vv_{B,i}; \vv_{D,i}]$ denotes the concatenation of $\vv_{B,i}$ and $\vv_{D,i}$.

\paragraph{Approach 2.} The second approach leverages a 1D convolutional neural network (1D-CNN), which processes data efficiently for large-scale graphs with high-dimensional topological embeddings \cite{kiranyaz20191, kiranyaz20211d}. The rationale behind using a 1D-CNN is that the embeddings  $\vv_{B,i}$ and $\vv_{D,i}$ are derived from inverse transform sampling, as detailed in Eq. (\ref{eq:inversionsampling}), resulting in \emph{sorted} births and deaths in vector coordinates. These ordered structures encode meaningful spatial relationships between adjacent values, analogous to how nearby pixels in images form edges or patches of color. In the embeddings $\vv_{B,i}$ and $\vv_{D,i}$, adjacent values represent local changes in the underlying topology, forming different stair shapes in Betti curves. Each stair shape corresponds to a distinct topological pattern, enabling the 1D-CNN to capture local dependencies and efficiently extract meaningful features, much like CNNs process image patches to detect edges and textures. Conceptually, by applying sliding convolutional filters to the concatenated topological embeddings, which are ordered and one-dimensional, the 1D-CNN efficiently captures hierarchical and spatial patterns:
$\vh_{\text{CNN}} = \text{CNN}_{\text{1D}}([\vv_{B,i}; \vv_{D,i}])$,
where $\text{CNN}_{\text{1D}}(\cdot)$ applies 1D convolutional layers with sliding filters followed by pooling operations, and $[\vv_{B,i}; \vv_{D,i}]$ is the concatenation of $\vv_{B,i}$ and $\vv_{D,i}$.

\subsection{Structural representation learning}

To effectively capture structural embeddings, GNNs are a suitable choice due to their ability to aggregate information from a node's neighbors using the graph's adjacency relationships. This aggregation process allows GNNs to encode both local and global structural patterns, ensuring that the node embeddings reflect not only their individual features but also the features and relationships of their neighboring nodes \cite{xu2018powerful, wu2020comprehensive}. As a result, GNNs are well-suited for learning the connectivity patterns inherent in structural graphs.

Recall that $\mX = (x_{ij})_{1\leq i \leq |V|,\;1 \leq j \leq d}$ represents the matrix of initial node features, where $|V|$ denotes the number of nodes in the structural graph $\mG_{struct}$, and $d$ denotes the dimensionality of the node feature vectors. Specifically, $\vx_i^\top \in \sR^d$ denotes the feature vector for node $i$. Initialize $\vh_v^{(1)} = \vx_v^\top$. GNNs update node embeddings through a series of graph convolutional layers:
$\vh_v^{(l+1)} = \text{GNNLayer}^{(l)}(\vh_v^{(l)}, \{\vh_u^{(l)} : u \in \sN(v)\})$,
where $\vh_v^{(l)}$ is the embedding of node $v$ at layer $l$, and $\sN(v)$ denotes the set of neighbors of node $v$.

After $L$ layers, a graph-level representation is computed by aggregating the final node embeddings to produce the graph representation:
$\vh_{G} = \text{Readout}(\{\vh_v^{(L)} : v \in \sV \})$,
where $\vh_G$ is the resulting graph-level embedding, $\sV$ is the set of all nodes in the graph, and Readout$(\cdot)$ represents an aggregation function. Common choices for Readout$(\cdot)$ include summation, mean, max pooling, or a learned attention mechanism over the node embeddings.

\subsection{Functional-structural integration}

There are various ways to integrate functional and structural representations. Here, we propose two options with distinct approaches. The first option involves concatenating the two representations
$\vh_{\text{fc\_struct}} = [\vh_G; \vh_{\text{MLP}}],$
which enables the model to learn interactions between functional and structural features. The second option involves elementwise addition of the two representations, provided their dimensionalities align
$\vh_{\text{fc\_struct}} = \vh_{\text{CNN}} + \vh_G,$
which offers a \emph{residual connection-like} integration \cite{he2016deep} that preserves the structural signals derived from the GNN and the functional signals provided by persistent graph homology.
For downstream tasks such as classification or regression, the integrated representation is used as input
$y=f(\vh_{\text{fc\_struct}}),$
where $f$ is typically an MLP or another classifier.

\newcommand{\boldmax}[1]{\textbf{#1}}

\begin{table*}
\customsmall
\setlength{\tabcolsep}{0.6mm}
\centering
\begin{tabular}{lcccccccccccccccccccc}
\toprule
& \multicolumn{2}{c}{DD} & \multicolumn{2}{c}{ENZYMES} & \multicolumn{2}{c}{BZR} & \multicolumn{2}{c}{COX2} & \multicolumn{2}{c}{MUTAG} & \multicolumn{2}{c}{NCI1} & \multicolumn{2}{c}{NCI109} & \multicolumn{2}{c}{PROTEINS} & \multicolumn{2}{c}{PROTEINS\_{f}} & \multicolumn{2}{c}{PTC\_MR} \\
Model & Acc & F1 & Acc & F1 & Acc & F1 & Acc & F1 & Acc & F1 & Acc & F1 & Acc & F1 & Acc & F1 & Acc & F1 & Acc & F1 \\
\midrule
FC\textsubscript{cnn}GCN & 76.32 & 76.77 & 35.11 & 35.54 & 81.98 & 80.87 & 78.73 & 73.65 & 80.84 & \boldmax{83.46} & 71.36 & 71.52 & 68.77 & 70.22 & \boldmax{75.23} & 73.57 & 74.69 & 73.75 & 54.16 & 49.42 \\
FC\textsubscript{mlp}GCN & \boldmax{78.18} & \boldmax{77.77} & \boldmax{40.17} & \boldmax{39.42} & \boldmax{83.37} & \boldmax{82.42} & \boldmax{81.80} & \boldmax{79.05} & \boldmax{81.04} & 82.51 & \boldmax{73.99} & \boldmax{74.34} & \boldmax{71.02} & \boldmax{71.19} & 74.84 & \boldmax{74.83} & \boldmax{75.17} & \boldmax{74.75} & 55.62 & 54.57\\
GCN & 70.38 & 70.97 & 27.39 & 26.24 & 81.32 & 78.24 & 80.80 & 76.75 & 72.38 & 73.21 & 68.71 & 68.39 & 67.24 & 68.11 & 69.66 & 67.66 & 69.00 & 68.45 & \boldmax{56.58} & \boldmax{54.93} \\
\midrule
FC\textsubscript{cnn}GIN & 77.59 & 76.29 & 39.72 & 40.45 & 81.98 & 81.50 & 80.44 & 78.64 & 81.75 & 83.93 & 75.57 & 76.54 & 74.32 & 74.54 & 74.75 & 73.47 & 74.72 & 73.60 & \boldmax{55.23} & 53.79 \\
FC\textsubscript{mlp}GIN & \boldmax{78.52} & \boldmax{78.31} & \boldmax{46.94} & \boldmax{46.13} & \boldmax{82.72} & \boldmax{81.69} & \boldmax{81.73} & 79.76 & 82.63 & 84.31 & \boldmax{78.30} & \boldmax{78.30} & \boldmax{75.53} & \boldmax{75.58} & \boldmax{75.08} & \boldmax{74.43} & \boldmax{75.41} & \boldmax{74.20} & 54.85 & \boldmax{53.97} \\
GIN & 71.48 & 70.31 & 36.50 & 38.24 & 82.06 & 80.25 & 81.73 & \boldmax{79.95} & \boldmax{83.34} & \boldmax{84.55} & 74.91 & 74.84 & 73.88 & 74.17 & 70.80 & 68.77 & 70.20 & 68.95 & 53.00 & 52.72 \\
\midrule
FC\textsubscript{cnn}GAT & 75.55 & 76.09 & 35.28 & 35.15 & 78.77 & 71.25 & 78.59 & 68.65 & \boldmax{81.74} & 82.34 & 64.91 & 61.06 & 63.58 & \boldmax{63.60} & 74.87 & 73.66 & 74.99 & 73.62 & 54.55 & 49.64 \\
FC\textsubscript{mlp}GAT & \boldmax{78.01} & \boldmax{77.30} & \boldmax{36.72} & \boldmax{36.78} & \boldmax{79.18} & \boldmax{72.68} & \boldmax{79.02} & \boldmax{69.70} & 81.74 & \boldmax{83.82} & \boldmax{65.09} & \boldmax{63.26} & \boldmax{64.50} & 62.83 & \boldmax{75.17} & \boldmax{74.43} & \boldmax{75.26} & \boldmax{74.61} & 53.68 & \boldmax{50.48} \\
GAT & 67.66 & 62.65 & 24.39 & 21.34 & 78.77 & 66.00 & 78.16 & 67.29 & 71.85 & 64.71 & 58.73 & 55.81 & 61.23 & 57.66 & 69.15 & 66.56 & 68.73 & 65.93 & \boldmax{57.65} & 50.25 \\
\midrule
FC\textsubscript{cnn}GSAGE & 77.76 & 77.38 & 36.67 & 38.16 & \boldmax{80.66} & 78.14 & 81.01 & 78.20 & 80.67 & 82.55 & 72.41 & 73.19 & 70.00 & 70.97 & 74.75 & 73.60 & \boldmax{75.20} & 73.67 & 56.58 & \boldmax{58.29} \\
FC\textsubscript{mlp}GSAGE & \boldmax{78.01} & \boldmax{78.03} & \boldmax{42.94} & \boldmax{42.36} & 80.41 & \boldmax{81.07} & \boldmax{83.15} & \boldmax{81.47} & \boldmax{81.03} & \boldmax{83.23} & \boldmax{74.79} & \boldmax{75.17} & \boldmax{72.45} & \boldmax{73.20} & \boldmax{75.08} & \boldmax{74.89} & 74.96 & \boldmax{74.77} & 55.72 & 55.80 \\
GSAGE & 69.44 & 70.34 & 30.44 & 28.77 & 79.92 & 77.31 & 82.44 & 80.92 & 74.50 & 74.98 & 70.27 & 70.02 & 69.01 & 68.88 & 69.03 & 67.49 & 69.21 & 67.22 & \boldmax{57.56} & 57.83 \\
\midrule
FC\textsubscript{cnn}ChebNet & \boldmax{79.52} & \boldmax{79.26} & 44.50 & 44.08 & 83.95 & 82.81 & 81.06 & 79.61 & \boldmax{85.26} & \boldmax{85.27} & 78.63 & 78.50 & 75.50 & 75.40 & 74.44 & 73.77 & 74.44 & 73.79 & \boldmax{59.42} & \boldmax{58.81} \\
FC\textsubscript{mlp}ChebNet & 79.10 & 78.95 & \boldmax{47.00} & \boldmax{46.30} & 83.46 & 82.70 & 82.98 & 82.19 & 82.11 & 81.70 & \boldmax{80.05} & \boldmax{80.00} & \boldmax{76.07} & \boldmax{75.98} & \boldmax{75.34} & \boldmax{74.98} & \boldmax{75.43} & \boldmax{75.06} & 59.42 & 58.51 \\
ChebNet & 73.45 & 72.51 & 38.00 & 37.06 & \boldmax{84.69} & \boldmax{83.19} & \boldmax{84.68} & \boldmax{83.72} & 80.00 & 79.48 & 77.13 & 77.08 & 74.70 & 74.65 & 70.49 & 69.74 & 70.22 & 69.53 & 57.10 & 56.88 \\
\midrule
FC\textsubscript{cnn}GATv2 & 77.82 & 77.27 & 36.67 & 35.87 & 81.23 & 75.33 & 78.09 & 70.11 & \boldmax{82.11} & \boldmax{81.31} & 64.52 & 64.49 & \boldmax{64.16} & \boldmax{63.67} & 74.62 & 74.03 & 74.71 & 74.09 & 58.84 & 57.10 \\
FC\textsubscript{mlp}GATv2 & \boldmax{79.94} & \boldmax{79.47} & \boldmax{38.00} & \boldmax{37.59} & \boldmax{83.46} & \boldmax{78.85} & \boldmax{79.36} & \boldmax{72.26} & 81.05 & 80.60 & \boldmax{65.33} & \boldmax{65.14} & 62.87 & 62.87 & \boldmax{75.07} & \boldmax{74.61} & \boldmax{75.16} & \boldmax{74.68} & 58.55 & \boldmax{57.45} \\
GATv2 & 69.07 & 64.33 & 22.83 & 18.59 & 80.00 & 71.13 & 78.30 & 68.92 & 69.47 & 63.94 & 60.50 & 58.89 & 61.62 & 59.21 & 67.00 & 63.54 & 67.00 & 63.64 & \boldmax{60.87} & 52.49 \\
\midrule
FC\textsubscript{cnn}ARMA & 78.81 & 78.54 & 42.17 & 41.25 & 84.69 & 81.31 & 80.43 & 77.82 & \boldmax{85.26} & \boldmax{85.17} & 76.07 & 76.03 & 73.77 & 73.68 & 74.80 & 74.10 & 74.53 & 73.85 & 59.71 & 59.11 \\
FC\textsubscript{mlp}ARMA & \boldmax{79.10} & \boldmax{78.89} & \boldmax{46.67} & \boldmax{46.07} & 84.69 & \boldmax{83.06} & 82.77 & \boldmax{81.10} & 84.74 & 84.75 & \boldmax{79.08} & \boldmax{78.96} & \boldmax{76.55} & \boldmax{76.52} & \boldmax{75.25} & \boldmax{74.90} & \boldmax{75.16} & \boldmax{74.81} & \boldmax{60.58} & \boldmax{60.19} \\
ARMA & 73.59 & 72.74 & 35.67 & 34.34 & \boldmax{84.94} & 82.40 & \boldmax{83.19} & 80.75 & 79.47 & 78.89 & 76.72 & 76.65 & 75.10 & 75.02 & 70.04 & 69.07 & 70.49 & 69.55 & 57.39 & 57.07 \\
\midrule
FC\textsubscript{cnn}TAG & \boldmax{79.52} & \boldmax{79.22} & 43.50 & 42.41 & 82.96 & 81.00 & 81.70 & 80.25 & \boldmax{84.21} & \boldmax{83.98} & 78.02 & 77.93 & 75.79 & 75.76 & 74.71 & 74.13 & 74.80 & 74.24 & \boldmax{61.16} & \boldmax{60.83} \\
FC\textsubscript{mlp}TAG & 79.24 & 79.10 & \boldmax{48.67} & \boldmax{47.84} & \boldmax{85.68} & \boldmax{84.30} & 83.62 & 81.96 & 82.63 & 82.01 & \boldmax{80.29} & \boldmax{80.28} & \boldmax{76.47} & \boldmax{76.46} & \boldmax{75.34} & \boldmax{74.99} & \boldmax{75.34} & \boldmax{74.99} & 58.55 & 57.86 \\
TAG & 72.88 & 72.21 & 37.00 & 36.12 & 85.19 & 84.05 & \boldmax{85.32} & \boldmax{84.31} & 79.47 & 78.69 & 76.40 & 76.36 & 75.18 & 75.11 & 70.13 & 68.77 & 70.13 & 68.77 & 58.84 & 58.60 \\
\midrule
FC\textsubscript{cnn}GraphConv & 78.53 & 78.29 & 43.67 & 43.04 & \boldmax{84.20} & 81.57 & 80.21 & 78.00 & 86.32 & 86.49 & 75.06 & 75.05 & 73.73 & 73.69 & 74.71 & 74.06 & 74.89 & 74.24 & 57.68 & 56.87 \\
FC\textsubscript{mlp}GraphConv & \boldmax{79.52} & \boldmax{79.42} & \boldmax{50.17} & \boldmax{49.78} & 84.20 & \boldmax{83.25} & 82.98 & 81.56 & 85.26 & 85.47 & \boldmax{78.22} & \boldmax{78.16} & \boldmax{75.38} & \boldmax{75.37} & \boldmax{75.52} & \boldmax{75.15} & \boldmax{75.52} & \boldmax{75.15} & \boldmax{58.55} & \boldmax{58.34} \\
GraphConv & 70.90 & 70.85 & 36.00 & 33.73 & 83.21 & 80.61 & \boldmax{84.04} & \boldmax{81.88} & \boldmax{86.84} & \boldmax{87.06} & 72.18 & 72.16 & 71.51 & 71.51 & 68.79 & 67.78 & 68.79 & 67.78 & 56.81 & 54.56 \\
\midrule
FC\textsubscript{cnn} & 75.21 & 73.21 & 28.89 & 27.49 & 78.77 & 72.68 & 78.16 & 68.92 & 81.76 & 80.56 & 62.63 & 61.48 & 60.33 & 58.41 & 73.61 & 72.79 & 73.58 & 72.77 & 54.66 & 51.25 \\
FC\textsubscript{mlp} & 75.64 & 74.81 & 29.11 & 27.04 & 78.52 & 71.55 & 78.16 & 68.92 & 82.09 & 80.52 & 62.48 & 61.44 & 60.87 & 59.37 & 74.12 & 73.52 & 74.48 & 73.52 & 53.59 & 52.07 \\
\bottomrule

\end{tabular}
\caption{Comparison of model performance across 10 biological datasets (DD, ENZYMES, BZR, COX2, MUTAG, NCI1, NCI109, PROTEINS, PROTEINS\_{full}, PTC\_MR), reported as average accuracy scores and average weighted F1 score.}
\label{table:results_core}
\end{table*}

\begin{table}[t]
\customsmallB
\setlength{\tabcolsep}{2mm}
\centering
\begin{tabular}{lcccccc}
\toprule
& \multicolumn{2}{c}{COLLAB} & \multicolumn{2}{c}{IMDB\_BI} & \multicolumn{2}{c}{IMDB\_MUL} \\
Model & Acc & F1 & Acc & F1 & Acc & F1 \\
\midrule
FC\textsubscript{cnn}GCN & \boldmax{76.30} & \boldmax{75.93} & \boldmax{68.50} & \boldmax{68.32} & \boldmax{48.13} & \boldmax{46.02} \\
FC\textsubscript{mlp}GCN & 73.70 & 73.61 & 67.80 & 67.45 & 45.27 & 43.61 \\
GCN & 57.97 & 62.78 & 48.80 & 56.67 & 33.07 & 37.49 \\
\midrule
FC\textsubscript{cnn}GIN & \boldmax{78.33} & \boldmax{78.15} & 68.50 & \boldmax{68.42} & \boldmax{48.80} & 46.84 \\
FC\textsubscript{mlp}GIN & 75.57 & 75.30 & \boldmax{68.60} & 68.39 & 48.73 & \boldmax{47.23} \\
GIN & 69.73 & 72.90 & 67.50 & 67.51 & 47.53 & 44.60 \\
\midrule
FC\textsubscript{cnn}GAT & \boldmax{75.97} & \boldmax{76.03} & \boldmax{69.40} & \boldmax{69.18} & \boldmax{47.73} & \boldmax{45.71} \\
FC\textsubscript{mlp}GAT & 72.87 & 72.86 & 67.60 & 67.10 & 45.27 & 43.26 \\
GAT & 52.03 & 67.68 & 49.20 & 60.29 & 33.33 & 35.96 \\
\midrule
FC\textsubscript{cnn}GSAGE & \boldmax{77.83} & \boldmax{77.59} & \boldmax{68.10} & 67.77 & \boldmax{48.13} & \boldmax{46.32} \\
FC\textsubscript{mlp}GSAGE & 73.73 & 73.63 & 67.90 & \boldmax{67.80} & 46.67 & 44.41 \\
GSAGE & 54.33 & 70.88 & 48.50 & 61.63 & 32.13 & 37.58 \\
\midrule
FC\textsubscript{cnn} & 76.53 & 76.43 & 69.80 & 69.09 & 48.47 & 46.45 \\
FC\textsubscript{mlp} & 71.70 & 71.80 & 67.90 & 67.75 & 46.13 & 43.35 \\
\bottomrule
\end{tabular}
\caption{Comparison of model performance across 3 social datasets (COLLAB, IMDB\_BINARY, IMDB\_MULTI), reported as average accuracy scores and average weighted F1 score.}
\label{table:results_core_social}
\end{table}

\begin{table*}[t]
\customsmallB
\centering
\begin{tabular}{lcccccccccc}
\toprule
& \multicolumn{2}{c}{COLLAB} & \multicolumn{2}{c}{IMDB\_BI} & \multicolumn{2}{c}{MUTAG} & \multicolumn{2}{c}{ENZYMES} & \multicolumn{2}{c}{PROTEINS} \\
Model & Acc & F1 & Acc & F1 & Acc & F1 & Acc & F1 & Acc & F1 \\
\midrule
GCN (None) & 57.88 & 59.32 & 47.94 & 35.76 & 74.25 & 73.09 & 26.45 & 23.32 & 69.62 & 68.73 \\
GCN + DIGL & 53.69 & 39.98 & 56.37 & 51.63 & 71.20 & 69.54 & 25.07 & 22.00 & 69.76 & 68.39 \\
GCN + SDRF & 64.13 & 62.23 & 48.58 & 34.19 & 71.85 & 70.17 & 24.35 & 21.20 & 70.12 & 68.79 \\
GCN + FoSR & 64.16 & 62.32 & 48.20 & 32.96 & 75.80 & 74.20 & 20.75 & 17.64 & 70.80 & 69.50 \\
GCN + PANDA & 66.28 & 64.92 & 61.96 & 61.63 & \boldmax{82.35} & \boldmax{81.80} & 27.77 & 26.86 & \boldmax{74.79} & \boldmax{74.27} \\
\midrule
FC\textsubscript{mlp}GCN & \boldmax{72.80} & \boldmax{72.73} & \boldmax{69.84} & \boldmax{69.38} & 79.70 & 79.47 & \boldmax{37.78} & \boldmax{37.10} & 73.25 & 72.69 \\
\midrule

GIN (None) & 69.50 & 71.30 & 69.62 & 69.51 & 82.40 & 82.28 & 33.90 & 32.89 & 70.09 & 69.37 \\
GIN + DIGL & 54.28 & 42.16 & 52.42 & 41.31 & 71.55 & 68.20 & 28.48 & 26.90 & 66.45 & 62.48 \\
GIN + SDRF & 63.29 & 58.09 & 52.77 & 40.52 & 73.45 & 70.91 & 29.48 & 27.52 & 67.79 & 64.76 \\
GIN + FoSR & 63.50 & 58.30 & 50.57 & 36.13 & 76.85 & 74.42 & 22.00 & 19.81 & 70.29 & 66.96 \\
GIN + PANDA & 69.93 & 69.24 & 66.54 & 66.24 & \boldmax{83.85} & \boldmax{83.48} & 34.52 & 34.01 & \boldmax{73.93} & \boldmax{73.43} \\
\midrule
FC\textsubscript{mlp}GIN & \boldmax{74.50} & \boldmax{74.26} & \boldmax{70.39} & \boldmax{70.00} & 82.65 & 82.44 & \boldmax{43.55} & \boldmax{42.73} & 73.03 & 72.68 \\
\bottomrule
\end{tabular}
\caption{Comparison of FC\textsubscript{mlp} with SOTA models under 80\%/10\%/10\% data split (Training/Validation/Testing) across 5 datasets (COLLAB, IMDB\_BINARY, MUTAG, ENZYMES, PROTEINS), reported as average accuracy and average weighted F1 scores.}
\label{table:results_sota}
\end{table*}

\begin{table*}[t]
\customsmallB
\centering
\begin{tabular}{lcccccccccc}
\toprule
& \multicolumn{2}{c}{COLLAB} & \multicolumn{2}{c}{IMDB\_BI} & \multicolumn{2}{c}{MUTAG} & \multicolumn{2}{c}{ENZYMES} & \multicolumn{2}{c}{PROTEINS} \\
Model & Acc & F1 & Acc & F1 & Acc & F1 & Acc & F1 & Acc & F1 \\
\midrule
GCN (None) & 52.80 & 41.85 & 50.00 & 37.33 & 64.70 & 61.27 & 20.52 & 17.09 & 65.51 & 63.90 \\
GCN + DIGL & 52.03 & 35.67 & 49.93 & 33.79 & 64.88 & 54.59 & 19.83 & 14.59 & 63.74 & 60.16 \\
GCN + SDRF & 61.34 & 59.12 & 49.65 & 32.94 & 65.28 & 55.53 & 19.32 & 13.90 & 63.86 & 60.09 \\
GCN + FoSR & 57.31 & 47.62 & 49.65 & 32.94 & 66.12 & 56.88 & 18.85 & 13.16 & 64.55 & 60.97 \\
GCN + PANDA & 63.33 & 62.27 & 56.89 & 54.67 & \boldmax{75.61} & \boldmax{73.51} & 21.38 & 19.30 & \boldmax{70.53} & \boldmax{69.59} \\
\midrule
FC\textsubscript{mlp}GCN & \boldmax{65.32} & \boldmax{65.14} & \boldmax{62.58} & \boldmax{61.94} & 71.84 & 70.35 & \boldmax{23.60} & \boldmax{22.03} & 68.10 & 67.80 \\
\midrule
GIN (None) & \boldmax{67.04} & 66.12 & \boldmax{65.53} & \boldmax{64.24} & 67.64 & 64.95 & 22.89 & 21.58 & 62.67 & 61.96 \\
GIN + DIGL & 53.81 & 41.81 & 54.23 & 44.94 & 62.95 & 54.53 & 22.00 & 19.60 & 63.23 & 57.96 \\
GIN + SDRF & 56.36 & 45.61 & 52.60 & 38.92 & 63.29 & 55.61 & 21.79 & 19.64 & 64.00 & 60.01 \\
GIN + FoSR & 57.34 & 48.12 & 51.85 & 37.50 & 66.82 & 61.28 & 20.34 & 17.91 & 67.14 & 63.29 \\
GIN + PANDA & 65.21 & 64.33 & 62.39 & 60.21 & \boldmax{73.81} & \boldmax{71.41} & 23.00 & 21.57 & \boldmax{68.20} & \boldmax{67.61} \\
\midrule
FC\textsubscript{mlp}GIN & 66.80 & \boldmax{66.46} & 62.97 & 62.39 & 70.56 & 69.14 & \boldmax{25.27} & \boldmax{23.76} & 66.99 & 66.66 \\
\bottomrule
\end{tabular}
\caption{Comparison of FC\textsubscript{mlp} with SOTA models under  5\%/5\%/90\% data split (Training/Validation/Testing).}
\label{table:results_sota_extra}
\end{table*}

\section{Applications in Graph-Level Classification}

\paragraph{Datasets.}
To rigorously assess cross-domain generalizability, we selected \textbf{13} datasets spanning two distinct domains: 
(1) Biological networks including ENZYMES, DD, NCI1, MUTAG, PTC\_MR, PROTEINS, PROTEINS\_full, NCI109, BZR, and COX2 \cite{morris2020tudataset}; and
(2) Social networks comprising COLLAB, IMDB-BINARY, and IMDB-MULTI \cite{morris2020tudataset}.
These datasets, which vary in graph size, density, class distribution, and domain characteristics, serve as standard benchmarks for evaluating the generalization and robustness of graph representation learning models. Their standardized use enables consistent comparisons across studies. Figure \ref{fig:schematic} presents a visualization of a representative example from MUTAG, along with Betti plots illustrating the average topological differences between samples corresponding to the two class labels in MUTAG, Class 0 and Class 1. Further dataset details are provided in the supplementary material.

\paragraph{Benchmarking models.}

We evaluated a comprehensive suite of models to assess graph-level classification performance. First, we examined \textbf{2} functional connectivity models: FC\textsubscript{CNN} (1D-CNN-based) and FC\textsubscript{MLP} (MLP-based). In addition, we assessed \textbf{9} GNN architectures: GCN \cite{kipf2016semi}, GIN \cite{xu2018powerful}, GAT \cite{velivckovic2017graph}, GSAGE \cite{hamilton2017inductive}, ChebNet \cite{defferrard2016chebnet}, GATv2 \cite{brody2021gatv2}, ARMAConv \cite{bianchi2021arma}, TAGConv \cite{du2017tagconv}, and GraphConv \cite{morris2019graphconv}. We further evaluated multi-modal approaches that integrate functional and structural connectivity by testing all combinations of functional connectivity learning (FC\textsubscript{CNN} and FC\textsubscript{MLP}) and structural GNNs, resulting in \textbf{18} Functional Connectivity Graph Neural Networks (e.g., FC\textsubscript{CNN}GCN, FC\textsubscript{MLP}GCN, FC\textsubscript{CNN}GIN, FC\textsubscript{MLP}GIN). Finally, we compared our proposed FC-GNNs to \textbf{4} state-of-the-art baseline methods: DIGL \cite{gasteiger2019digl}, SDRF \cite{topping2021sdrf}, FoSR \cite{karhadkar2022fosr}, and PANDA \cite{choi2024panda}. Additional experimental details, analyses, results, and full code are provided in the supplementary material.

\paragraph{Evaluation protocols.} For within-architecture comparisons among FC-GNNs and GNN baselines, we employed nested cross-validation (CV) for robust performance estimates: the inner loop used stratified 5-fold CV for hyperparameter tuning, and the outer loop used stratified 5-fold CV for generalization assessment, with stratification preserving class label proportions in each fold. Grid search was conducted over key hyperparameters, including hidden dimensions, learning rate, weight decay, number of epochs, and loss function. For comparisons with state-of-the-art baseline methods, we adapted the protocol used in prior work \cite{choi2024panda}, averaging model performance over 100 random data splits and reporting mean test accuracy and weighted F1 score. Hyperparameter settings, code, and further details are provided in the supplementary material.

\subsection{Results and interpretation}

Tables~\ref{table:results_core} and \ref{table:results_core_social} present within-architecture comparison results for integrating functional connectivity modules with 2-layer GNNs. Dual-modality FC-GNNs consistently outperform their single-modality counterparts across diverse datasets. Multi-modal integration yields an average \textbf{5.33}\% accuracy improvement and \textbf{1.68}\% variance reduction across 10 biological and 3 social network datasets, indicating that functional and structural connectivity provide complementary learning signals. While 2-layer GNNs capture local structural patterns, increasing depth enables aggregation of more complex, higher-order relationships. To demonstrate that these benefits are not limited to shallow models, we also conducted experiments with 3-layer GNN architectures; the results similarly show that functional connectivity continues to provide complementary signals as the structural backbone becomes more expressive. Full results for 3-layer architectures are provided in the appendix tables in the supplementary material.

Table~\ref{table:results_sota} presents comparisons with state-of-the-art baseline methods, showing that our FC\textsubscript{mlp}GCN and FC\textsubscript{mlp}GIN models achieve competitive or superior performance across benchmark datasets. In addition, we also evaluate our methods under limited supervision (5\%/5\%/90\% train/val/test split), as shown in Table~\ref{table:results_sota_extra}, where it consistently outperforms the baselines, highlighting the data efficiency and strong generalization of our approach. This improvement is especially pronounced for social networks. For example, COLLAB, with an average of 2,457 edges per graph, presents highly dense connectivity where traditional message passing suffers from over-smoothing. IMDB-BINARY, which lacks node features, is generally limited to extracting graph representations from structural topology alone. The proposed FC-GNN enables explicit extraction of functional topology, providing a crucial inductive bias for these structure-dependent tasks and capturing global graph properties that complement local message passing. On molecular datasets, FC-GNN achieves strong results, notably outperforming baselines on the challenging ENZYMES task, which highlights its ability to capture complex biochemical patterns. For MUTAG and PROTEINS, slightly lower performance compared to the state-of-the-art PANDA can be attributed to their simpler structures and distinctive chemical motifs, where PANDA's targeted strategies are particularly effective.
Overall, the FC-GNN dual-pipeline combines local message passing with global topological features, consistently improving performance across structurally diverse network datasets.

\section*{Broader Impact} 
Our framework provides a foundation for multi-modal graph analysis by linking functional interactions with network structure of complex systems. It enables insights in domains where functional and structural connectivity offer complementary perspectives, capturing interactions across both local and global structure and contributing to emerging directions in geometric deep learning.

\bibliographystyle{plainnat}
\bibliography{reference}

\clearpage
\appendix

\section{Pipeline Implementation}
\subsection{Baseline Methods}
We selected 9 representative GNN architectures that cover a broad spectrum of design paradigms—spectral (GCN \cite{kipf2016semi}, ChebNet \cite{defferrard2016chebnet}, and ARMAConv \cite{bianchi2021arma}), attention-based (GAT \cite{velivckovic2017graph} and GATv2 \cite{brody2021gatv2}), inductive sampling (GraphSAGE \cite{hamilton2017inductive}), topology-adaptive filtering (TAGConv \cite{du2017tagconv}), and higher-order message passing (GIN \cite{xu2018powerful} and GraphConv \cite{morris2019graphconv}). These models are widely used in the graph classification literature, and their diversity contributes to the coverage of the major families of GNN approaches. To ensure consistent training pipelines and fair comparisons, we implemented all baselines using PyTorch Geometric (PyG) \cite{fey2019fast}. Detailed model implementations are provided in \texttt{topo\_concat.py} and \texttt{topo\_only.py}.

In addition, we compare against 4 state-of-the-art graph classification baseline methods:

\paragraph{DIGL.} Diffusion Improves Graph Learning (DIGL) \cite{gasteiger2019digl} is a diffusion-based rewiring method using kernel smoothing followed by edge sparsification.

\paragraph{SDRF.} Stochastic Discrete Ricci Flow (SDRF) \cite{topping2021sdrf} rewires edges with low Ricci curvature to alleviate bottlenecks; We evaluate the \textit{original variant} (edge additions only). 

\paragraph{FoSR.} First-order Spectral Rewiring (FoSR) \cite{karhadkar2022fosr} rewires based on iterative spectral gap maximization to prevent over-squashing.

\paragraph{PANDA.} Expanded width-aware message passing paradigm \cite{choi2024panda}
expands message passing where high-centrality nodes receive larger hidden dimensions to preserve signal flow. \\

\subsection{Functional Connectivity GNNs}
We provide implementation details of the Functional Connectivity Block (FCB) pipeline in \texttt{topo\_concat.py}. The preprocessing step is implemented in \texttt{preprocess.py}, which extracts topological features for each dataset prior to training. We hereby prove the stability of the topological features after inverse transform sampling upsample or downsample the persistent graph homology (PGH) representation into a consistent dimensionality across graphs of varying sizes.

\paragraph{Proposition A.2} \label{prop:stability}
A collection of vectors $\{\vv_{B,i}\}_{i=1}^N$, equipped with  the $p$-norm metric $d_{p,B}(\vv_{B,i},\vv_{B,j}) = ||\vv_{B,i}-\vv_{B,j}||_p$, defines a metric space such that:
    \begin{align*} 
        \lim_{m\to\infty} \frac{d_{p,B}}{m} &= \Big(\int_0^1 |F^{-1}_{\mG_{fc}^{(i)},B}(z) - F^{-1}_{\mG_{fc}^{(j)},B}(z)|^p\,dz\Big)^{\frac{1}{p}} \\
        &= W_{p,B},
    \end{align*}
where $W_{p,B}$ is the $p$-Wasserstein distance between persistence diagrams of connected components. Similarly, for a vector space representing cycles $\{\vv_{D,i}\}_{i=1}^N$, equipped  with the $p$-norm metric $d_{p,D}$, we have  $\lim_{n\to\infty} \frac{d_{p,D}}{n} = W_{p,D}$

\paragraph{Proof.}
Recall that $m \in \{1,2,3,...\}$. Without loss of generality, we will prove the claim for persistence diagrams of connected components, as the case of persistence diagrams of cycles follows from an analogous argument.
\begin{alignat*}{2}
    &\frac{d_{p,B}(\vv_{B,i},\vv_{B,j})}{m} \\
    &= \frac{||\vv_{B,i}-\vv_{B,j}||_p}{m} \\
    &= \frac{1}{m} \Big(\sum_{k=1}^m | F^{-1}_{\mG_{fc}^{(i)},B}(\frac{k}{m}) - F^{-1}_{\mG_{fc}^{(j)},B}(\frac{k}{m}) |^p\Big)^{\frac{1}{p}} \\
    &= \Big( \frac{1}{m^p} \sum_{k=1}^m | F^{-1}_{\mG_{fc}^{(i)},B}(\frac{k}{m}) - F^{-1}_{\mG_{fc}^{(j)},B}(\frac{k}{m}) |^p\Big)^{\frac{1}{p}} \\
    &= \Big( \sum_{k=1}^m \Big| \frac{F^{-1}_{\mG_{fc}^{(i)},B}(\frac{k}{m}) - F^{-1}_{\mG_{fc}^{(j)},B}(\frac{k}{m})}{m} \Big|^p\Big)^{\frac{1}{p}}.
\end{alignat*}
Therefore,
\begin{alignat*}{2}
    &\lim_{m\to\infty} \frac{d_{p,B}(\vv_{B,i},\vv_{B,j})}{m} \\
    &= \lim_{m\to\infty} \Big( \sum_{k=1}^m \Big| \frac{F^{-1}_{\mG_{fc}^{(i)},B}(\frac{k}{m}) - F^{-1}_{\mG_{fc}^{(j)},B}(\frac{k}{m})}{m} \Big|^p\Big)^{\frac{1}{p}} \\
    &= \Big(\int_0^1 |F^{-1}_{\mG_{fc}^{(i)},B}(z) - F^{-1}_{\mG_{fc}^{(j)},B}(z)|^p\,dz\Big)^{\frac{1}{p}}.
\end{alignat*}

\paragraph{Remark.}
Proposition \ref{prop:stability} demonstrates that the $p$-norm distance in the specified vector spaces corresponds to the $p$-Wasserstein distance in the original space of persistence diagrams. As a result, the central stability theorem \cite{skraba2020wasserstein} applies to these vector spaces, implying that the embedding of functional connectivity remains stable under perturbations in the inputs.

\section{Dataset}

Table~\ref{tab:dataset_statistics} provides a summary of the key statistics for the 13 datasets used in our experiments. We provide a brief description of each dataset:
\paragraph{PROTEINS} \cite{borgwardt2005protein}: A dataset representing protein structures where nodes correspond to secondary structure elements and edges denote spatial proximity.
\paragraph{ENZYMES} \cite{schomburg2004brenda}: A dataset consists of protein tertiary structures classified into six enzyme classes.
\paragraph{DD} \cite{dobson2003distinguishing}: A dataset represents protein structures classified as enzymes or non-enzymes.
\paragraph{NCI1} \cite{wale2008comparison}: A dataset of chemical compounds screened for activity against non-small cell lung cancer and leukemia.
\paragraph{MUTAG} \cite{debnath1991structure}: A dataset of mutagenic aromatic and heteroaromatic compounds.
\paragraph{PTC\_MR} \cite{helma2001predictive}: A dataset for predictive toxicology in rodents.
\paragraph{PROTEINS\_full} \cite{borgwardt2005protein}: The full version of the PROTEINS dataset, containing additional annotations.
\paragraph{NCI109} \cite{wale2008comparison}: A chemical compound dataset similar to NCI1 but with a different screening focus.
\paragraph{BZR} \cite{sutherland2003spline}: A dataset of benzodiazepine derivatives for classification tasks.
\paragraph{COX2} \cite{sutherland2003spline}: A dataset of cyclooxygenase-2 inhibitors for classification tasks.
\paragraph{COLLAB} \cite{yanardag2015social}: A dataset of scientific collaboration networks where nodes represent researchers and edges denote co-authorships. 
\paragraph{IMDB-BINARY} \cite{yanardag2015social}: A dataset of ego-networks from the IMDB database. Each graph corresponds to an actor/actress, with edges indicating co-appearance in a movie. The task is to classify the movie genre into Action or Romance.
\paragraph{IMDB-MULTI} \cite{yanardag2015social}: Similar to IMDB-BINARY, the task is to classify the movie genre into Comedy, Romance, or Sci-Fi.

\begin{table}[ht]
\centering
\fontsize{9}{9}\selectfont
\setlength{\tabcolsep}{1mm}
\begin{tabular}{lcccc}
\hline
\textbf{Dataset} & \textbf{\#Graphs} & \textbf{\#Classes} & \textbf{Avg. Nodes} & \textbf{Avg. Edges} \\
\hline
PROTEINS         & 1,113     & 2   & 39.06 & 72.82 \\
ENZYMES          & 600       & 6   & 32.63 & 62.14 \\
DD               & 1,178     & 2   & 284.32& 715.66\\
NCI1             & 4,110     & 2   & 29.87 & 32.30 \\
MUTAG            & 188       & 2   & 17.93 & 19.79 \\
PTC\_MR          & 344       & 2   & 14.29 & 14.69 \\
PROTEINS\_full   & 1,113     & 2   & 39.06 & 72.82 \\
NCI109           & 4,127     & 2   & 29.68 & 32.13 \\
BZR              & 405       & 2   & 35.75 & 38.36 \\
COX2             & 467       & 2   & 41.22 & 43.45 \\
COLLAB           & 5,000     & 3   & 74.49 & 2457.78 \\
IMDB-BINARY      & 1,000     & 2   & 19.77 & 96.53 \\
IMDB-MULTI       & 1,500     & 3   & 13.00 & 65.94 \\
\hline
\end{tabular}
\caption{Dataset statistics for the selected TU datasets.}
\label{tab:dataset_statistics}
\end{table}

These datasets are widely used in the context of graph classification tasks and are directly obtained from TUDatasets \cite{morris2020tudataset}. Detailed implementation of dataset loading is provided in \texttt{dataset.py} and \texttt{helper.py}.

\section{Experimental Details}
\subsection{Hyperparameter Tuning and Evaluation Protocol}
For within-architecture comparisons among FC-GNNs and GNN baselines, we performed hyperparameter tuning to identify the best-performing configuration for each model. Our searching process focused on hidden dimension, learning rate, and weight decay rate, with the final selection determined through nested cross-validation (CV). For most datasets, we conducted three independent nested CV runs and reported the averaged result to ensure robustness against variance. For the large datasets (DD, NCI1, and NCI109), full repetitions were computationally infeasible: each trial required a single GPU (with 256GB RAM) for up to a week in our high-performance computing (HPC) environment. In these cases, we performed a single nested CV run. The nested CV design helps mitigate variance by averaging across folds, so even single-trial results remain reasonably reliable. Table~\ref{tab:hyperparameters_table} shows the search range of hyperparameters. Detailed implementation of nested CV and hyperparameter tuning is provided in \texttt{nested\_cv.py}, \texttt{train\_evaluate.py}, and \texttt{main.py}.

\begin{table}[ht]
\centering
\setlength{\tabcolsep}{1mm}
\begin{tabular}{lc}
\hline
\textbf{Hyperparameters} & \textbf{Search Space} \\
\hline
Hidden Dimension      & 32, 64, 128 \\
Learning Rate (lr)    & $1 \times 10^{-5}$, $1 \times 10^{-4}$, $1 \times 10^{-3}$ \\
Weight Decay          & $1 \times 10^{-5}$, $1 \times 10^{-4}$, $1 \times 10^{-3}$ \\
Number of Epochs      & 200 \\
Criterion             & CrossEntropyLoss \\
\hline
\end{tabular}
\caption{Summary of the hyperparameter configurations used in the within-architecture experiments.}
\label{tab:hyperparameters_table}
\end{table}

For comparisons with state-of-the-art baseline methods, we adapted the protocol used in prior work \cite{choi2024panda}. All models are implemented on top of GCN \cite{kipf2016semi} and GIN \cite{xu2018powerful} backbones for consistency. For FC-GNN, we applied an additional MLP layer after the convolution blocks to incorporate topological signals. To ensure a fair and reproducible evaluation, we followed a standardized 100-run protocol using random data splits with an 80\%/10\%/10\% train/validation/test split. For each trial, the configuration at the final epoch was selected. We reported the test accuracy and weighted-F1 score for each run, and summarized performance by computing the mean and standard deviation across the 100 trials. Table~\ref{tab:sota_common_hyper} shows the fixed training hyperparameters across all methods to ensure fairness. Table~\ref{tab:sota_method_specific} shows the specific configuration for each baseline method.

\begin{table}[ht]
\centering
\setlength{\tabcolsep}{1mm}
\begin{tabular}{lc}
\hline
\textbf{Hyperparameters} & \textbf{Value} \\
\hline
Number of GNN layers  & 2 \\
Dropout rate          & 0.5 \\
Pooling Function      & Global Mean \\
Hidden Dimension      & 64 \\
Learning Rate (lr)    & $1 \times 10^{-3}$ \\
Weight Decay          & $1 \times 10^{-5}$ \\
Optimizer             & Adam \\
Activation Function   & ReLU \\
Batch Size            & 64 \\
Number of Epochs      & 200 \\
Criterion             & CrossEntropyLoss \\
\hline
\end{tabular}
\caption{Common hyperparameter configuration for SOTA comparison experiments. Note that for PANDA and other rewiring methods, num\_layers=1 is passed since their architectures internally instantiate 2 convolution layers. Dropout placement differs across methods: rewiring methods apply dropout between convolution layers, whereas FC-GNN applies it after pooling.}
\label{tab:sota_common_hyper}
\end{table}

\begin{table}[ht]
\centering
\setlength{\tabcolsep}{1mm}
\begin{tabular}{lc}
\hline
\textbf{Method} & \textbf{Additional Hyperparameters} \\
\hline
DIGL  & \makecell[l]{- Iterations: 10 \\ - Diffusion coefficient ($\alpha$): 0.1 \\ - Sparsification threshold ($\varepsilon$): 0.05 \\ - Number of relations: 2} \\
\hline
SDRF  & \makecell[l]{- Ricci flow iterations: 10 \\ - Number of relations: 2} \\
\hline
FoSR  & \makecell[l]{- Spectral rewiring iterations: 10 \\ - Number of relations: 2} \\
\hline
PANDA & \makecell[l]{- Top-k nodes: 7 \\ - Expanded dimension ($p_{high}$): 128 \\ - Centrality measure: degree \\ - Expansion factor: 2 \\ - Number of relations: 2} \\
\hline
FC-GNN & \makecell[l]{- Topological feature dropout: 0.3} \\
\hline
\end{tabular}
\caption{Method-specific hyperparameters for SOTA comparison experiments. Note that the topological feature dropout is specific to FCGNN's internal architecture and not shared across baseline methods.}
\label{tab:sota_method_specific}
\end{table}

\subsection{Hardware Specifications and Libraries}

All experiments were conducted on an HPC cluster. Each node was equipped with NVIDIA RTX 2080 Ti GPUs (11GB VRAM) and NVIDIA RTX A5000 GPUs (24GB VRAM).

For within-architecture comparisons among FC-GNNs and GNN baselines, we used a conda environment with Python~3.12.9, CUDA~12.1, PyTorch~2.5.1, PyTorch Geometric~2.6.1, NumPy~2.0.1, and NetworkX~3.4.2. For state-of-the-art baseline comparisons, 
we followed the open-source implementation of PANDA \cite{choi2024panda} using a separate environment with Python~3.11.3, CUDA~12.1, PyTorch~2.7.1, PyTorch Geometric~2.6.1, NumPy~2.2.6, and NetworkX~3.5. A full list of dependencies is provided in  \texttt{requirements\_core.txt} and \texttt{requirements\_sota.txt}.

\section{Within-architecture Comparisons}
\begin{figure*}[ht]
    \centering
    \includegraphics[width=\textwidth]{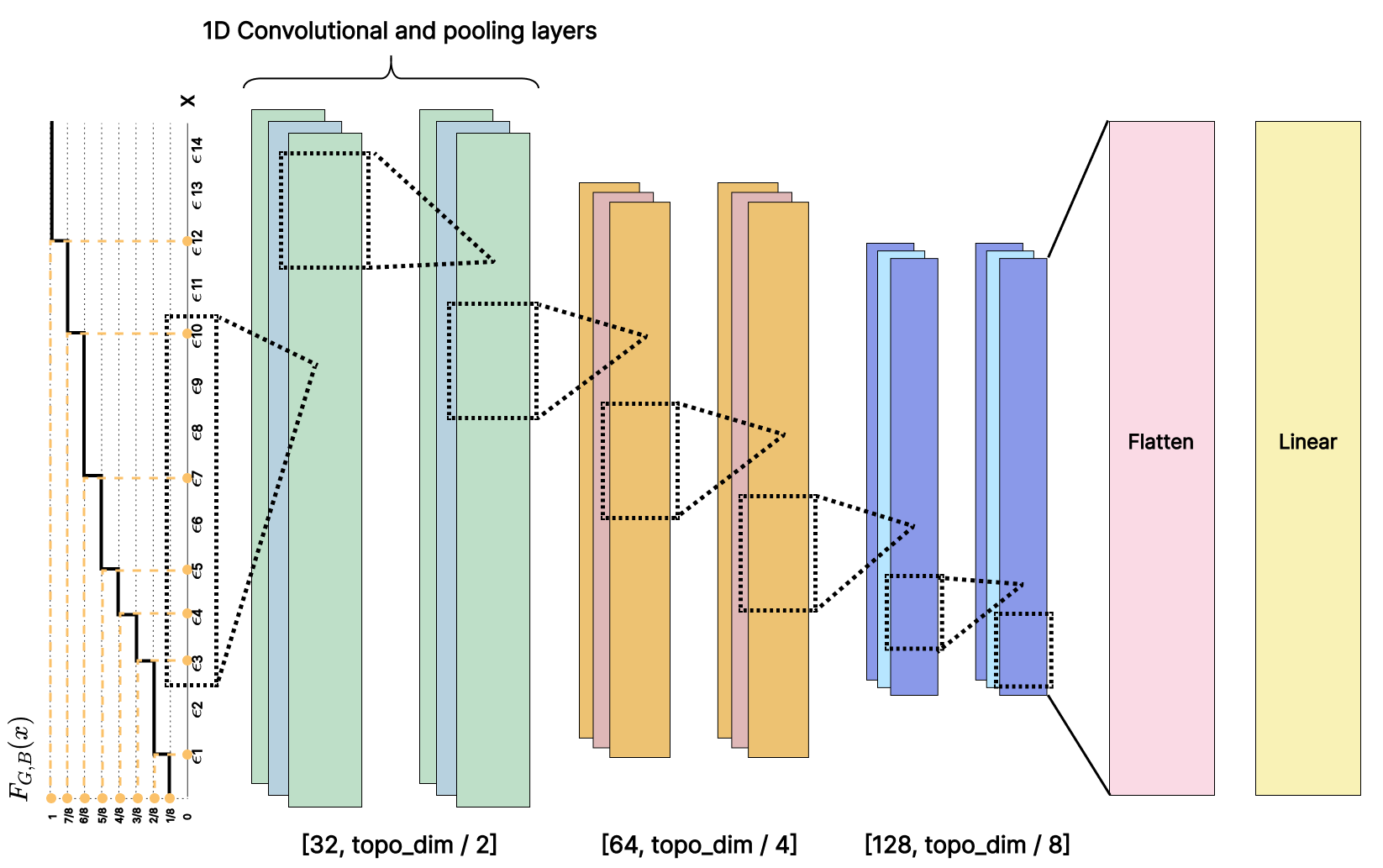}
    \caption{The figure illustrates the processing pipeline of topological features derived from inverse transform sampling results. The input, represented as an embedding vector, undergoes a series of 1D convolutional and pooling layers to extract meaningful patterns. Each convolutional block progressively reduces the dimensionality of the topological feature space, transforming it through layers with different channel configurations, where topo\_dim represents the original topological feature dimension. Finally, the extracted features are flattened and passed through a linear layer for downstream classification tasks. The dotted lines highlight the hierarchical feature extraction process across different resolutions.}
    \label{fig:cnn}
\end{figure*}

In our experiments, we implement two variants of the Functional Connectivity Block (FCB): \textbf{FC\textsubscript{CNN}} and \textbf{FC\textsubscript{MLP}}. \textbf{FC\textsubscript{CNN}} applies three consecutive 1D convolutional layers 
(kernel size 5, padding 2), each followed by ReLU and MaxPool1d for downsampling.  The resulting feature maps are flattened and passed through a fully connected layer with ReLU and a dropout of 0.3, projecting the output into the hidden dimension 
(see Figure~\ref{fig:cnn}). \textbf{FC\textsubscript{MLP}} processes topological features using two fully connected layers, each followed by ReLU and a dropout of 0.3. In both variants, the transformed topological embedding is concatenated with the GNN embedding, and an additional linear layer unifies the representation back into the hidden dimension.

\subsection{Experiments using 2-layer architectures}
We used 2-layer GNNs as baseline models to evaluate structural GNN performance. These baselines help to assess whether multi-modal approaches that integrate functional and structural connectivity via the Functional Connectivity Block (FCB) provide additional benefits. In addition, we include results for standalone FCB models to highlight their independent effectiveness.

\Cref{tab:2layer-all-4backbone-acc,tab:2layer-all-4backbone-f1,tab:2layer-12-5backbone-acc,tab:2layer-12-5backbone-f1} summarizes the performance of 2-layer baselines and their FC-GNN counterparts across benchmark datasets. Note that the COLLAB dataset was not evaluated on the five additional backbones (ChebNet, GATv2, TAGConv, GraphConv, ARMA) because training did not finish within a runtime threshold of one week per model on our HPC cluster. For fairness and reproducibility, we therefore report COLLAB results only for the four common backbones (GCN, GIN, GAT, GraphSAGE), while including all other datasets for the extended backbone experiments. 

\paragraph{Results.} 
Across benchmark datasets, FC\textsubscript{mlp} variants consistently outperform their 2-layer GNN baselines, with particularly large gains on complex multi-class datasets and social networks. On ENZYMES, FC\textsubscript{mlp}GIN improves accuracy 
from 36.50\% to 46.94\%, representing a 10.44 percentage points improvement, while COLLAB increases from 57.97\% (GCN) to 76.30\% (FC\textsubscript{mlp}GCN), which is an 18.33 point increase, and IMDB-BINARY rises from 48.80\% (GCN) to 68.50\% (FC\textsubscript{mlp}GCN), achieving an improvement of 19.70 points. Notably, even standalone FC modules (FC\textsubscript{cnn}, FC\textsubscript{mlp}) achieve competitive performance, underscoring the independent value of functional connectivity features. Weighted F1 scores confirm these findings, showing balanced improvements across classes, with ENZYMES improving by +7.89 points from GIN to FC\textsubscript{mlp}GIN, and consistent gains on feature-sparse social networks. Extending to 12 datasets and 
five additional GNN backbones, FC\textsubscript{mlp} variants again demonstrate architecture-agnostic improvements. For example, FC\textsubscript{mlp}GraphConv achieves 50.17\% on ENZYMES versus 36.00\% for GraphConv (+14.17 points), while the weighted F1 score for IMDB-MULTI improves from 33.94 to 46.36 (+12.42 points). These comprehensive results confirm that functional connectivity augmentation delivers balanced classification improvements across all evaluated architectures, with particularly strong gains on multi-class problems and challenging social network datasets.

\begin{table*}[ht]
\footnotesize
\centering
\setlength{\tabcolsep}{0.8mm}
\begin{tabular}{lccccccc}
\toprule
Model & DD & ENZYMES & BZR & COX2 & MUTAG & NCI1 & NCI109 \\
\midrule
FC\textsubscript{cnn}GCN & 76.32$\pm$2.65 & 35.11$\pm$2.79 & 81.98$\pm$2.37 & 78.73$\pm$2.74 & 80.84$\pm$4.92 & 71.36$\pm$1.51 & 68.77$\pm$0.68 \\
FC\textsubscript{mlp}GCN & \boldmax{78.18}$\pm$1.47 & \boldmax{40.17}$\pm$3.61 & \boldmax{83.37}$\pm$2.25 & \boldmax{81.80}$\pm$2.47 & \boldmax{81.04}$\pm$4.63 & \boldmax{73.99}$\pm$1.99 & \boldmax{71.02}$\pm$1.53 \\
GCN & 70.38$\pm$2.68 & 27.39$\pm$2.40 & 81.32$\pm$1.85 & 80.80$\pm$2.24 & 72.38$\pm$3.85 & 68.71$\pm$1.59 & 67.24$\pm$0.92 \\
\midrule
FC\textsubscript{cnn}GIN & 77.59$\pm$2.91 & 39.72$\pm$3.23 & 81.98$\pm$2.86 & 80.44$\pm$2.93 & 81.75$\pm$5.28 & 75.57$\pm$0.46 & 74.32$\pm$1.10 \\
FC\textsubscript{mlp}GIN & \boldmax{78.52}$\pm$1.39 & \boldmax{46.94}$\pm$3.08 & \boldmax{82.72}$\pm$2.79 & \boldmax{81.73}$\pm$2.09 & 82.63$\pm$5.73 & \boldmax{78.30}$\pm$0.37 & \boldmax{75.53}$\pm$1.30 \\
GIN & 71.48$\pm$3.76 & 36.50$\pm$4.15 & 82.06$\pm$3.01 & 81.73$\pm$1.79 & \boldmax{83.34}$\pm$6.90 & 74.91$\pm$0.76 & 73.88$\pm$1.23 \\
\midrule
FC\textsubscript{cnn}GAT & 75.55$\pm$1.45 & 35.28$\pm$3.32 & 78.77$\pm$2.47 & 78.59$\pm$0.70 & \boldmax{81.74}$\pm$3.91 & 64.91$\pm$1.51 & 63.58$\pm$1.66 \\
FC\textsubscript{mlp}GAT & \boldmax{78.01}$\pm$0.71 & \boldmax{36.72}$\pm$2.38 & \boldmax{79.18}$\pm$2.30 & \boldmax{79.02}$\pm$1.08 & 81.74$\pm$4.50 & \boldmax{65.09}$\pm$1.63 & \boldmax{64.50}$\pm$2.12 \\
GAT & 67.66$\pm$0.74 & 24.39$\pm$2.32 & 78.77$\pm$0.49 & 78.16$\pm$0.41 & 71.85$\pm$4.47 & 58.73$\pm$4.05 & 61.23$\pm$1.73 \\
\midrule
FC\textsubscript{cnn}GSAGE & 77.76$\pm$1.31 & 36.67$\pm$3.05 & \boldmax{80.66}$\pm$2.17 & 81.01$\pm$2.89 & 80.67$\pm$4.54 & 72.41$\pm$1.32 & 70.00$\pm$1.38 \\
FC\textsubscript{mlp}GSAGE & \boldmax{78.01}$\pm$1.50 & \boldmax{42.94}$\pm$1.91 & 80.41$\pm$3.20 & \boldmax{83.15}$\pm$1.55 & \boldmax{81.03}$\pm$4.55 & \boldmax{74.79}$\pm$1.65 & \boldmax{72.45}$\pm$1.55 \\
GSAGE & 69.44$\pm$2.90 & 30.44$\pm$4.21 & 79.92$\pm$2.57 & 82.44$\pm$1.85 & 74.50$\pm$3.91 & 70.27$\pm$0.88 & 69.01$\pm$0.92 \\
\midrule
FC\textsubscript{cnn} & 75.21$\pm$2.09 & 28.89$\pm$2.70 & 78.77$\pm$0.49 & 78.16$\pm$0.41 & 81.76$\pm$4.96 & 62.63$\pm$1.98 & 60.33$\pm$1.22 \\
FC\textsubscript{mlp} & 75.64$\pm$0.78 & 29.11$\pm$1.99 & 78.52$\pm$0.99 & 78.16$\pm$0.41 & 82.09$\pm$5.18 & 62.48$\pm$1.67 & 60.87$\pm$1.84 \\
\bottomrule

\toprule
Model & PROTEINS & PROTEINS\_{full} & PTC\_MR & COLLAB & IMDB\_BINARY & IMDB\_MULTI & / \\
\midrule
FC\textsubscript{cnn}GCN & \boldmax{75.23}$\pm$1.87 & 74.69$\pm$1.09 & 54.16$\pm$5.24 & \boldmax{76.30}$\pm$0.59 & \boldmax{68.50}$\pm$0.71 & \boldmax{48.13}$\pm$1.28 & / \\
FC\textsubscript{mlp}GCN & 74.84$\pm$1.84 & \boldmax{75.17}$\pm$1.54 & 55.62$\pm$4.12 & 73.70$\pm$0.64 & 67.80$\pm$1.75 & 45.27$\pm$2.07& / \\
GCN & 69.66$\pm$2.48 & 69.00$\pm$2.05 & \boldmax{56.58}$\pm$3.69 & 57.97$\pm$0.95 & 48.80$\pm$3.75 & 33.07$\pm$1.37 & / \\
\midrule
FC\textsubscript{cnn}GIN & 74.75$\pm$1.76 & 74.72$\pm$1.73 & \boldmax{55.23}$\pm$3.71 & \boldmax{78.33}$\pm$0.34 & 68.50$\pm$1.00 & \boldmax{48.80}$\pm$2.18 & / \\
FC\textsubscript{mlp}GIN & \boldmax{75.08}$\pm$2.06 & \boldmax{75.41}$\pm$1.85 & 54.85$\pm$2.90 & 75.57$\pm$0.47 & \boldmax{68.60}$\pm$1.32 & 48.73$\pm$1.90 & / \\
GIN & 70.80$\pm$1.27 & 70.20$\pm$1.95 & 53.00$\pm$4.36 & 69.73$\pm$0.87 & 67.50$\pm$2.93 & 47.53$\pm$1.83 & / \\
\midrule
FC\textsubscript{cnn}GAT & 74.87$\pm$1.61 & 74.99$\pm$1.44 & 54.55$\pm$4.03 & \boldmax{75.97}$\pm$0.62 & \boldmax{69.40}$\pm$2.08 & \boldmax{47.73}$\pm$2.90 & / \\
FC\textsubscript{mlp}GAT & \boldmax{75.17}$\pm$1.70 & \boldmax{75.26}$\pm$2.06 & 53.68$\pm$5.75 & 72.87$\pm$0.37 & 67.60$\pm$1.07 & 45.27$\pm$0.65 & / \\
GAT & 69.15$\pm$2.36 & 68.73$\pm$1.79 & \boldmax{57.65}$\pm$3.38 & 52.03$\pm$0.87 & 49.20$\pm$4.28 & 33.33$\pm$1.38 & / \\
\midrule
FC\textsubscript{cnn}GSAGE & 74.75$\pm$1.72 & \boldmax{75.20}$\pm$1.44 & 56.58$\pm$6.21 & \boldmax{77.83}$\pm$0.66 & \boldmax{68.10}$\pm$1.91 & \boldmax{48.13}$\pm$3.49 & / \\
FC\textsubscript{mlp}GSAGE & \boldmax{75.08}$\pm$2.16 & 74.96$\pm$1.96 & 55.72$\pm$7.01 & 73.73$\pm$0.09 & 67.90$\pm$1.39 & 46.67$\pm$2.83 & / \\
GSAGE & 69.03$\pm$1.82 & 69.21$\pm$2.38 & \boldmax{57.56}$\pm$5.54 & 54.33$\pm$1.46 & 48.50$\pm$4.87 & 32.13$\pm$1.80 & / \\
\midrule
FC\textsubscript{cnn} & 73.61$\pm$1.93 & 73.58$\pm$1.72 & 54.66$\pm$3.69 & 76.53$\pm$0.38 & 69.80$\pm$1.60 & \boldmax{48.47}$\pm$2.01 & / \\
FC\textsubscript{mlp} & 74.12$\pm$1.95 & 74.48$\pm$1.95 & 53.59$\pm$4.24 & 71.70$\pm$1.28 & 67.90$\pm$2.35 & 46.13$\pm$3.22 & / \\
\bottomrule
\end{tabular}
\caption{Comparison of model performance across 13 datasets (DD, ENZYMES, BZR, COX2, MUTAG, NCI1, NCI109, PROTEINS, PROTEINS\_{full}, PTC\_MR, COLLAB, IMDB\_BINARY, IMDB\_MULTI), reported as average accuracy scores$\pm$standard deviation.}
\label{tab:2layer-all-4backbone-acc}
\end{table*}

\begin{table*}[ht]
\footnotesize
\centering
\setlength{\tabcolsep}{0.8mm}
\begin{tabular}{lccccccc}
\toprule
Model & DD & ENZYMES & BZR & COX2 & MUTAG & NCI1 & NCI109 \\
\midrule
FC\textsubscript{cnn}GCN & 76.77$\pm$1.80 & 35.54$\pm$4.68 & 80.87$\pm$3.37 & 73.65$\pm$3.62 & \boldmax{83.46}$\pm$4.51 & 71.52$\pm$1.76 & 70.22$\pm$1.46\\
FC\textsubscript{mlp}GCN & \boldmax{77.77}$\pm$1.41 & \boldmax{39.42}$\pm$4.93 & \boldmax{82.42}$\pm$3.12 & \boldmax{79.05}$\pm$3.84 & 82.51$\pm$4.45 & \boldmax{74.34}$\pm$1.51 & \boldmax{71.19}$\pm$1.54\\
GCN & 70.97$\pm$3.08 & 26.24$\pm$4.28 & 78.24$\pm$3.37 & 76.75$\pm$3.55 & 73.21$\pm$5.83 & 68.39$\pm$2.20 & 68.11$\pm$1.72\\
\midrule
FC\textsubscript{cnn}GIN & 76.29$\pm$2.70 & 40.45$\pm$5.04 & 81.50$\pm$2.71 & 78.64$\pm$4.24 & 83.93$\pm$4.52 & 76.54$\pm$1.57 & 74.54$\pm$1.34\\
FC\textsubscript{mlp}GIN & \boldmax{78.31}$\pm$1.34 & \boldmax{46.13}$\pm$5.61 & \boldmax{81.69}$\pm$3.37 & 79.76$\pm$3.85 & 84.31$\pm$4.89 & \boldmax{78.30}$\pm$1.65 & \boldmax{75.58}$\pm$1.26\\
GIN & 70.31$\pm$2.62 & 38.24$\pm$4.23 & 80.25$\pm$3.09 & \boldmax{79.95}$\pm$4.39 & \boldmax{84.55}$\pm$4.17 & 74.84$\pm$2.32 & 74.17$\pm$1.69\\
\midrule
FC\textsubscript{cnn}GAT & 76.09$\pm$2.47 & 35.15$\pm$5.20 & 71.25$\pm$1.97 & 68.65$\pm$1.93 & 82.34$\pm$5.09 & 61.06$\pm$8.18 & \boldmax{63.60}$\pm$1.79\\
FC\textsubscript{mlp}GAT & \boldmax{77.30}$\pm$2.82 & \boldmax{36.78}$\pm$4.49 & \boldmax{72.68}$\pm$3.96 & \boldmax{69.70}$\pm$2.38 & \boldmax{83.82}$\pm$4.43 & \boldmax{63.26}$\pm$6.62 & 62.83$\pm$3.86\\
GAT & 62.65$\pm$5.58 & 21.34$\pm$4.38 & 66.00$\pm$14.84 & 67.29$\pm$4.24 & 64.71$\pm$5.65 & 55.81$\pm$9.22 & 57.66$\pm$6.99 \\
\midrule
FC\textsubscript{cnn}GSAGE & 77.38$\pm$1.50 & 38.16$\pm$4.47 & 78.14$\pm$3.30 & 78.20$\pm$3.71 & 82.55$\pm$4.39 & 73.19$\pm$1.81 & 70.97$\pm$1.31\\
FC\textsubscript{mlp}GSAGE & \boldmax{78.03}$\pm$1.34 & \boldmax{42.36}$\pm$4.80 & \boldmax{81.07}$\pm$3.12 & \boldmax{81.47}$\pm$3.07 & \boldmax{83.23}$\pm$5.01 & \boldmax{75.17}$\pm$1.50 & \boldmax{73.20}$\pm$1.22 \\
GSAGE & 70.34$\pm$3.15 & 28.77$\pm$3.39 & 77.31$\pm$3.71 & 80.92$\pm$4.04 & 74.98$\pm$6.09 & 70.02$\pm$2.07 & 68.88$\pm$1.89 \\
\midrule
FC\textsubscript{cnn} & 73.21$\pm$1.06 & 27.49$\pm$2.20 & 72.68$\pm$2.53 & 68.92$\pm$8.92 & 80.56$\pm$8.24 & 61.48$\pm$1.50 & 58.41$\pm$1.44 \\
FC\textsubscript{mlp} & 74.81$\pm$1.01 & 27.04$\pm$3.50 & 71.55$\pm$3.96 & 68.92$\pm$8.92 & 80.52$\pm$7.58 & 61.44$\pm$1.55 & 59.37$\pm$1.78 \\
\bottomrule
\toprule
Model & PROTEINS & PROTEINS\_full & PTC\_MR & COLLAB & IMDB\_BINARY & IMDB\_MULTI & / \\
\midrule
FC\textsubscript{cnn}GCN & 73.57$\pm$3.05 & 73.75$\pm$3.03 & 49.42$\pm$6.36 & \boldmax{75.93}$\pm$0.73 & \boldmax{68.32}$\pm$0.98 & \boldmax{46.02}$\pm$1.42 & / \\
FC\textsubscript{mlp}GCN & \boldmax{74.83}$\pm$3.21 & \boldmax{74.75}$\pm$2.83 & 54.57$\pm$6.46 & 73.61$\pm$0.65 & 67.45$\pm$1.74 & 43.61$\pm$2.92 & / \\
GCN & 67.66$\pm$2.89 & 68.45$\pm$2.64 & \boldmax{54.93}$\pm$5.01 & 62.78$\pm$0.67 & 56.67$\pm$3.52 & 37.49$\pm$1.37 & / \\
\midrule
FC\textsubscript{cnn}GIN & 73.47$\pm$2.88 & 73.60$\pm$2.93 & 53.79$\pm$5.95 & \boldmax{78.15}$\pm$0.51 & \boldmax{68.42}$\pm$1.02 & 46.84$\pm$1.91 & / \\
FC\textsubscript{mlp}GIN & \boldmax{74.43}$\pm$2.16 & \boldmax{74.20}$\pm$2.18 & \boldmax{53.97}$\pm$6.92 & 75.30$\pm$0.51 & 68.39$\pm$1.34 & \boldmax{47.23}$\pm$2.28 & / \\
GIN & 68.77$\pm$2.46 & 68.95$\pm$2.35 & 52.72$\pm$4.84 & 72.90$\pm$1.52 & 67.51$\pm$2.96 & 44.60$\pm$2.39 & / \\
\midrule
FC\textsubscript{cnn}GAT & 73.66$\pm$3.13 & 73.62$\pm$3.20 & 49.64$\pm$6.24 & \boldmax{76.03}$\pm$0.55 & \boldmax{69.18}$\pm$2.24 & \boldmax{45.71}$\pm$3.03 & / \\
FC\textsubscript{mlp}GAT & \boldmax{74.43}$\pm$3.14 & \boldmax{74.61}$\pm$2.86 & \boldmax{50.48}$\pm$6.01 & 72.86$\pm$0.08 & 67.10$\pm$1.26 & 43.26$\pm$1.49 & / \\
GAT & 66.56$\pm$3.51 & 65.93$\pm$3.48 & 50.25$\pm$3.81 & 67.68$\pm$0.79 & 60.29$\pm$6.23 & 35.96$\pm$3.77 & / \\
\midrule
FC\textsubscript{cnn}GSAGE & 73.60$\pm$3.36 & 73.67$\pm$3.19 & \boldmax{58.29}$\pm$6.33 & \boldmax{77.59}$\pm$0.71 & 67.77$\pm$1.67 & \boldmax{46.32}$\pm$3.76 & / \\
FC\textsubscript{mlp}GSAGE & \boldmax{74.89}$\pm$2.93 & \boldmax{74.77}$\pm$3.03 & 55.80$\pm$5.91 & 73.63$\pm$0.36 & \boldmax{67.80}$\pm$1.41 & 44.41$\pm$3.20 & / \\
GSAGE & 67.49$\pm$2.47 & 67.22$\pm$3.18 & 57.83$\pm$5.42 & 70.88$\pm$0.61 & 61.63$\pm$2.99 & 37.58$\pm$5.67 & / \\
\midrule
FC\textsubscript{cnn} & 72.79$\pm$2.30 & 72.77$\pm$2.31 & 51.25$\pm$6.24 & 76.43$\pm$0.42 & 69.09$\pm$1.50 & \boldmax{46.45}$\pm$2.13 & / \\
FC\textsubscript{mlp} & 73.52$\pm$2.34 & 73.52$\pm$2.34 & 52.07$\pm$5.53 & 71.80$\pm$1.36 & 67.75$\pm$2.29 & 43.35$\pm$4.20 & / \\
\bottomrule
\end{tabular}
\caption{Comparison of model performance across 13 datasets (DD, ENZYMES, BZR, COX2, MUTAG, NCI1, NCI109, PROTEINS, PROTEINS\_{full}, PTC\_MR, COLLAB, IMDB\_BINARY, IMDB\_MULTI), reported as average weighted F1 scores$\pm$standard deviation.}
\label{tab:2layer-all-4backbone-f1}
\end{table*}

\begin{table*}[t]
\footnotesize
\centering
\begin{tabular}{lcccccc}
\toprule
Model & DD & ENZYMES & BZR & COX2 & MUTAG & PTC\_MR \\
\midrule
FC\textsubscript{cnn}ChebNet & \boldmax{79.52}$\pm$1.97 & 44.50$\pm$5.21 & 83.95$\pm$3.74 & 81.06$\pm$5.96 & \boldmax{85.26}$\pm$3.16 & \boldmax{59.42}$\pm$3.04 \\
FC\textsubscript{mlp}ChebNet & 79.10$\pm$2.30 & \boldmax{47.00}$\pm$3.86 & 83.46$\pm$2.88 & 82.98$\pm$3.16 & 82.11$\pm$6.09 & 59.42$\pm$6.55 \\
ChebNet & 73.45$\pm$1.00 & 38.00$\pm$2.39 & \boldmax{84.69}$\pm$2.98 & \boldmax{84.68}$\pm$4.54 & 80.00$\pm$4.88 & 57.10$\pm$3.38 \\
\midrule
FC\textsubscript{cnn}GATv2 & 77.82$\pm$1.71 & 36.67$\pm$3.21 & 81.23$\pm$2.26 & 78.09$\pm$6.78 & \boldmax{82.11}$\pm$5.62 & 58.84$\pm$6.12 \\
FC\textsubscript{mlp}GATv2 & \boldmax{79.94}$\pm$1.91 & \boldmax{38.00}$\pm$2.67 & \boldmax{83.46}$\pm$2.54 & \boldmax{79.36}$\pm$7.27 & 81.05$\pm$6.32 & 58.55$\pm$3.85 \\
GATv2 & 69.07$\pm$4.52 & 22.83$\pm$4.96 & 80.00$\pm$2.64 & 78.30$\pm$6.54 & 69.47$\pm$8.25 & \boldmax{60.87}$\pm$3.67 \\
\midrule
FC\textsubscript{cnn}ARMA & 78.81$\pm$2.16 & 42.17$\pm$3.86 & 84.69$\pm$1.48 & 80.43$\pm$5.41 & \boldmax{85.26}$\pm$2.68 & 59.71$\pm$2.96 \\
FC\textsubscript{mlp}ARMA & \boldmax{79.10}$\pm$0.72 & \boldmax{46.67}$\pm$3.98 & 84.69$\pm$2.77 & 82.77$\pm$5.48 & 84.74$\pm$4.82 & \boldmax{60.58}$\pm$1.69 \\
ARMA & 73.59$\pm$1.22 & 35.67$\pm$2.81 & \boldmax{84.94}$\pm$2.64 & \boldmax{83.19}$\pm$5.72 & 79.47$\pm$5.37 & 57.39$\pm$1.48\\
\midrule
FC\textsubscript{cnn}TAG & \boldmax{79.52}$\pm$1.78 & 43.50$\pm$4.93 & 82.96$\pm$2.52 & 81.70$\pm$4.43 & \boldmax{84.21}$\pm$3.72 & \boldmax{61.16}$\pm$2.49 \\
FC\textsubscript{mlp}TAG & 79.24$\pm$1.80 & \boldmax{48.67}$\pm$4.96 & \boldmax{85.68}$\pm$3.18 & 83.62$\pm$4.23 & 82.63$\pm$7.74 & 58.55$\pm$5.07 \\
TAG & 72.88$\pm$0.92 & 37.00$\pm$3.23 & 85.19$\pm$1.91 & \boldmax{85.32}$\pm$4.92 & 79.47$\pm$5.86 & 58.84$\pm$4.36 \\
\midrule
FC\textsubscript{cnn}GraphConv & 78.53$\pm$1.64 & 43.67$\pm$2.77 & \boldmax{84.20}$\pm$1.44 & 80.21$\pm$3.90 & 86.32$\pm$5.10 & 57.68$\pm$4.14 \\
FC\textsubscript{mlp}GraphConv & \boldmax{79.52}$\pm$1.71 & \boldmax{50.17}$\pm$4.87 & 84.20$\pm$2.39 & 82.98$\pm$4.41 & 85.26$\pm$5.67 & \boldmax{58.55}$\pm$1.97 \\
GraphConv & 70.90$\pm$1.22 & 36.00$\pm$5.85 & 83.21$\pm$1.67 & \boldmax{84.04}$\pm$4.80 & \boldmax{86.84}$\pm$4.40 & 56.81$\pm$7.30\\
\midrule
FC\textsubscript{cnn} & 75.21$\pm$2.09 & 28.89$\pm$2.70 & 78.77$\pm$0.49 & 78.16$\pm$0.41 & 81.76$\pm$4.96 & 54.66$\pm$3.69 \\
FC\textsubscript{mlp} & 75.64$\pm$0.78 & 29.11$\pm$1.99 & 78.52$\pm$0.99 & 78.16$\pm$0.41 & 82.09$\pm$5.18 & 53.59$\pm$4.24\\
\bottomrule
\toprule
Model & NCI1 & NCI109 & PROTEINS & PROTEINS\_full & IMDB\_BINARY & IMDB\_MULTI\\
\midrule
FC\textsubscript{cnn}ChebNet & 78.63$\pm$1.01 & 75.50$\pm$1.95 & 74.44$\pm$1.86 & 74.44$\pm$2.29  & \boldmax{69.80}$\pm$1.44 & \boldmax{47.73}$\pm$1.29\\
FC\textsubscript{mlp}ChebNet & \boldmax{80.05}$\pm$1.46 & \boldmax{76.07}$\pm$0.69 & \boldmax{75.34}$\pm$2.60 & \boldmax{75.43}$\pm$2.65 & 68.80$\pm$0.75 & 44.47$\pm$2.89\\
ChebNet & 77.13$\pm$1.90 & 74.70$\pm$1.10 & 70.49$\pm$2.71 & 70.22$\pm$2.06 & 50.60$\pm$0.37 & 34.80$\pm$0.62\\
\midrule
FC\textsubscript{cnn}GATv2 & 64.52$\pm$0.75 & \boldmax{64.16}$\pm$2.49 & 74.62$\pm$2.00 & 74.71$\pm$2.74 & 69.10$\pm$1.66 & \boldmax{48.80}$\pm$2.26\\
FC\textsubscript{mlp}GATv2 & \boldmax{65.33}$\pm$2.19 & 62.87$\pm$1.93 & \boldmax{75.07}$\pm$1.94 & \boldmax{75.16}$\pm$2.00 & \boldmax{71.60}$\pm$3.01 & 43.00$\pm$2.51\\
GATv2 & 60.50$\pm$3.66 & 61.62$\pm$5.08 & 67.00$\pm$1.12 & 67.00$\pm$1.29 & 50.20$\pm$1.96 & 32.67$\pm$1.81\\
\midrule
FC\textsubscript{cnn}ARMA & 76.07$\pm$1.97 & 73.77$\pm$1.84 & 74.80$\pm$2.65 & 74.53$\pm$2.33 & 70.00$\pm$1.41 & \boldmax{46.27}$\pm$2.82\\
FC\textsubscript{mlp}ARMA & \boldmax{79.08}$\pm$1.90 & \boldmax{76.55}$\pm$1.57 & \boldmax{75.25}$\pm$2.87 & \boldmax{75.16}$\pm$2.87 & \boldmax{70.30}$\pm$3.47 & 45.27$\pm$2.88\\
ARMA & 76.72$\pm$0.92 & 75.10$\pm$0.35 & 70.04$\pm$3.09 & 70.49$\pm$2.81  & 53.30$\pm$2.38 & 39.07$\pm$3.80\\
\midrule
FC\textsubscript{cnn}TAG & 78.02$\pm$2.35 & 75.79$\pm$1.38 & 74.71$\pm$1.76 & 74.80$\pm$1.66  & \boldmax{70.20}$\pm$1.29 & \boldmax{45.53}$\pm$2.75\\
FC\textsubscript{mlp}TAG & \boldmax{80.29}$\pm$2.82 & \boldmax{76.47}$\pm$1.03 & \boldmax{75.34}$\pm$2.06 & \boldmax{75.34}$\pm$2.06  & 69.60$\pm$2.11 & 44.40$\pm$2.30\\
TAG & 76.40$\pm$0.85 & 75.18$\pm$1.04 & 70.13$\pm$2.17 & 70.13$\pm$2.17 & 50.20$\pm$2.01 & 35.40$\pm$2.53 \\
\midrule
FC\textsubscript{cnn}GraphConv & 75.06$\pm$1.86 & 73.73$\pm$1.46 & 74.71$\pm$2.26 & 74.89$\pm$2.32  & 70.50$\pm$0.71 & 44.73$\pm$2.33\\
FC\textsubscript{mlp}GraphConv & \boldmax{78.22}$\pm$0.81 & \boldmax{75.38}$\pm$1.19 & \boldmax{75.52}$\pm$2.38 & \boldmax{75.52}$\pm$2.38  & \boldmax{71.40}$\pm$2.91 & \boldmax{48.33}$\pm$0.73\\
GraphConv & 72.18$\pm$1.57 & 71.51$\pm$1.57 & 68.79$\pm$2.67 & 68.79$\pm$2.67 & 61.90$\pm$2.96 & 38.20$\pm$2.82 \\
\midrule
FC\textsubscript{cnn} & 62.63$\pm$1.98 & 60.33$\pm$1.22 & 73.61$\pm$1.93 & 73.58$\pm$1.72 & 69.80$\pm$1.60 & \boldmax{48.47}$\pm$2.01  \\
FC\textsubscript{mlp} & 62.48$\pm$1.67 & 60.87$\pm$1.84 & 74.12$\pm$1.95 & 74.48$\pm$1.95 & 67.90$\pm$2.35 & 46.13$\pm$3.22\\
\bottomrule
\end{tabular}
\caption{Comparison of model performance across 12 datasets (DD, ENZYMES, BZR, COX2, MUTAG, NCI1, NCI109, PROTEINS, PROTEINS\_{full}, PTC\_MR, IMDB\_BINARY, IMDB\_MULTI), reported as average accuracy scores$\pm$standard deviation.}
\label{tab:2layer-12-5backbone-acc}
\end{table*}

\begin{table*}[t]
\footnotesize
\centering
\begin{tabular}{lcccccc}
\toprule
Model & DD & ENZYMES & BZR & COX2 & MUTAG & PTC\_MR\\
\midrule
FC\textsubscript{cnn}ChebNet & \boldmax{79.26}$\pm$1.97 & 44.08$\pm$5.00 & 82.81$\pm$3.48 & 79.61$\pm$6.61 & \boldmax{85.27}$\pm$3.05 & \boldmax{58.81}$\pm$2.80\\
FC\textsubscript{mlp}ChebNet & 78.95$\pm$2.16 & \boldmax{46.30}$\pm$3.36 & 82.70$\pm$2.89 & 82.19$\pm$3.06 & 81.70$\pm$5.82 & 58.51$\pm$5.93 \\
ChebNet & 72.51$\pm$1.19 & 37.06$\pm$2.10 & \boldmax{83.19}$\pm$2.20 & \boldmax{83.72}$\pm$4.92 & 79.48$\pm$5.07 & 56.88$\pm$3.17 \\
\midrule
FC\textsubscript{cnn}GATv2 & 77.27$\pm$1.79 & 35.87$\pm$3.84 & 75.33$\pm$3.01 & 70.11$\pm$9.94 & \boldmax{81.31}$\pm$5.95 & 57.10$\pm$6.80 \\
FC\textsubscript{mlp}GATv2 & \boldmax{79.47}$\pm$1.88 & \boldmax{37.59}$\pm$3.39 & \boldmax{78.85}$\pm$3.68 & \boldmax{72.26}$\pm$10.02 & 80.60$\pm$6.59 & \boldmax{57.45}$\pm$4.44\\
GATv2 & 64.33$\pm$6.32 & 18.59$\pm$4.61 & 71.13$\pm$3.65 & 68.92$\pm$8.92 & 63.94$\pm$10.65 & 52.49$\pm$3.47 \\
\midrule
FC\textsubscript{cnn}ARMA & 78.54$\pm$2.26 & 41.25$\pm$2.87 & 81.31$\pm$3.40 & 77.82$\pm$6.92 & \boldmax{85.17}$\pm$2.22 & 59.11$\pm$3.36 \\
FC\textsubscript{mlp}ARMA & \boldmax{78.89}$\pm$0.88 & \boldmax{46.07}$\pm$3.55 & \boldmax{83.06}$\pm$2.79 & \boldmax{81.10}$\pm$6.72 & 84.75$\pm$4.76 & \boldmax{60.19}$\pm$1.65 \\
ARMA & 72.74$\pm$1.94 & 34.34$\pm$2.58 & 82.40$\pm$4.45 & 80.75$\pm$7.26 & 78.89$\pm$5.69 & 57.07$\pm$0.99 \\
\midrule
FC\textsubscript{cnn}TAG & \boldmax{79.22}$\pm$1.89 & 42.41$\pm$4.19 & 81.00$\pm$2.93 & 80.25$\pm$5.08 & \boldmax{83.98}$\pm$3.81 & \boldmax{60.83}$\pm$2.24 \\
FC\textsubscript{mlp}TAG & 79.10$\pm$1.94 & \boldmax{47.84}$\pm$4.85 & \boldmax{84.30}$\pm$2.57 & 81.96$\pm$4.72 & 82.01$\pm$8.01  & 57.86$\pm$4.93 \\
TAG & 72.21$\pm$1.86 & 36.12$\pm$3.59 & 84.05$\pm$1.84 & \boldmax{84.31}$\pm$5.46 & 78.69$\pm$5.80 & 58.60$\pm$3.96\\
\midrule
FC\textsubscript{cnn}GraphConv & 78.29$\pm$1.66 & 43.04$\pm$2.79 & 81.57$\pm$2.01 & 78.00$\pm$4.93 & 86.49$\pm$4.86 & 56.87$\pm$4.46\\
FC\textsubscript{mlp}GraphConv & \boldmax{79.42}$\pm$1.81 & \boldmax{49.78}$\pm$4.75 & \boldmax{83.25}$\pm$2.82 & 81.56$\pm$5.03 & 85.47$\pm$5.43 & \boldmax{58.34}$\pm$2.15\\
GraphConv & 70.85$\pm$1.30 & 33.73$\pm$6.29 & 80.61$\pm$2.66 & \boldmax{81.88}$\pm$5.71 & \boldmax{87.06}$\pm$4.23 & 54.56$\pm$4.97 \\
\midrule
FC\textsubscript{cnn} & 73.21$\pm$1.06 & 27.49$\pm$2.20 & 72.68$\pm$2.53 & 68.92$\pm$8.92 & 80.56$\pm$8.24 & 51.25$\pm$6.24 \\
FC\textsubscript{mlp} & 74.81$\pm$1.01 & 27.04$\pm$3.50 & 71.55$\pm$3.96 & 68.92$\pm$8.92 & 80.52$\pm$7.58 & 52.07$\pm$5.53 \\
\bottomrule
\toprule
Model & NCI1 & NCI109 & PROTEINS & PROTEINS\_full & IMDB\_BINARY & IMDB\_MULTI\\
\midrule
FC\textsubscript{cnn}ChebNet & 78.50$\pm$1.03 & 75.40$\pm$2.05 & 73.77$\pm$2.11 & 73.79$\pm$2.62  & \boldmax{69.21}$\pm$1.86 & \boldmax{45.06}$\pm$1.48\\
FC\textsubscript{mlp}ChebNet & \boldmax{80.00}$\pm$1.47 & \boldmax{75.98}$\pm$0.75 & \boldmax{74.98}$\pm$2.80 & \boldmax{75.06}$\pm$2.84  & 67.84$\pm$1.43 & 40.90$\pm$3.66\\
ChebNet & 77.08$\pm$1.89 & 74.65$\pm$1.14 & 69.74$\pm$2.58 & 69.53$\pm$1.77 & 41.30$\pm$5.88 & 26.68$\pm$2.77\\

\midrule
FC\textsubscript{cnn}GATv2 & 64.49$\pm$0.71 & \boldmax{63.67}$\pm$3.08 & 74.03$\pm$2.02 & 74.09$\pm$3.06  & 68.52$\pm$1.92 & \boldmax{46.89}$\pm$2.31\\
FC\textsubscript{mlp}GATv2 & \boldmax{65.14}$\pm$2.15 & 62.87$\pm$1.93 & \boldmax{74.61}$\pm$2.12 & \boldmax{74.68}$\pm$2.17   & \boldmax{70.95}$\pm$2.87 & 39.57$\pm$2.92\\
GATv2 & 58.89$\pm$4.85 & 59.21$\pm$7.38 & 63.54$\pm$1.56 & 63.64$\pm$1.24 & 38.08$\pm$3.05 & 25.13$\pm$2.07\\
\midrule
FC\textsubscript{cnn}ARMA & 76.03$\pm$2.00 & 73.68$\pm$1.83 & 74.10$\pm$2.95 & 73.85$\pm$2.66  & \boldmax{69.77}$\pm$1.58 & \boldmax{43.79}$\pm$3.29\\
FC\textsubscript{mlp}ARMA & \boldmax{78.96}$\pm$1.97 & \boldmax{76.52}$\pm$1.55 & \boldmax{74.90}$\pm$3.05 & \boldmax{74.81}$\pm$3.04  & 69.61$\pm$3.77 & 42.50$\pm$3.19\\
ARMA & 76.65$\pm$0.91 & 75.02$\pm$0.44 & 69.07$\pm$3.65 & 69.55$\pm$2.75  & 48.48$\pm$5.19 & 35.16$\pm$4.85\\

\midrule
FC\textsubscript{cnn}TAG & 77.93$\pm$2.43 & 75.76$\pm$1.39 & 74.13$\pm$2.04 & 74.24$\pm$1.91  & \boldmax{69.99}$\pm$1.34 & \boldmax{42.57}$\pm$3.76\\
FC\textsubscript{mlp}TAG & \boldmax{80.28}$\pm$2.82 & \boldmax{76.46}$\pm$1.05 & \boldmax{74.99}$\pm$2.22 & \boldmax{74.99}$\pm$2.22 & 68.97$\pm$2.26 & 41.15$\pm$3.00\\
TAG & 76.36$\pm$0.83 & 75.11$\pm$1.08 & 68.77$\pm$2.22 & 68.77$\pm$2.22  & 43.64$\pm$3.39 & 28.75$\pm$4.20\\
\midrule
FC\textsubscript{cnn}GraphConv & 75.05$\pm$1.86 & 73.69$\pm$1.45 & 74.06$\pm$2.57 & 74.24$\pm$2.62  & 70.24$\pm$0.62 & 41.62$\pm$3.58\\
FC\textsubscript{mlp}GraphConv & \boldmax{78.16}$\pm$0.78 & \boldmax{75.37}$\pm$1.20 & \boldmax{75.15}$\pm$2.59 & \boldmax{75.15}$\pm$2.59  & \boldmax{71.08}$\pm$3.04 & \boldmax{46.36}$\pm$1.97\\
GraphConv & 72.16$\pm$1.59 & 71.51$\pm$1.57 & 67.78$\pm$2.90 & 67.78$\pm$2.90  & 61.23$\pm$3.54 & 33.94$\pm$2.89\\
\midrule
FC\textsubscript{cnn} & 61.48$\pm$1.50 & 58.41$\pm$1.44 & 72.79$\pm$2.30 & 72.77$\pm$2.31  & 69.09$\pm$1.50 & \boldmax{46.45}$\pm$2.13\\
FC\textsubscript{mlp} & 61.44$\pm$1.55 & 59.37$\pm$1.78 & 73.52$\pm$2.34 & 73.52$\pm$2.34  & 67.75$\pm$2.29 & 43.35$\pm$4.20\\
\bottomrule
\end{tabular}
\caption{Comparison of model performance across 12 datasets (DD, ENZYMES, BZR, COX2, MUTAG, NCI1, NCI109, PROTEINS, PROTEINS\_{full}, PTC\_MR, IMDB\_BINARY, IMDB\_MULTI), reported as average weighted F1 scores$\pm$standard deviation.}
\label{tab:2layer-12-5backbone-f1}
\end{table*}

\subsection{Experiments using 3-layer architectures}

To further isolate the effect of the FC block, we conducted an additional set of experiments using 3-layer GNN backbones, extending the 2-layer architectures with one additional GNN layer. We selected 6 representative datasets—ENZYMES, BZR, COX2, MUTAG, PROTEINS, and IMDB-BINARY—chosen to span both biological domains (biochemical and protein classification) and social networks, as well as to cover a range of task complexities from relatively simple binary classification to more challenging multi-class problems. This design allows us to evaluate the robustness of FC augmentation across diverse graph learning scenarios.

The results in Tables~\ref{tab:3layer-6dataset-acc} and \ref{tab:3layer-6dataset-f1} show that FC-augmented models consistently outperform their plain 3-layer counterparts in both accuracy and weighted F1 scores. The most notable gains appear on ENZYMES and PROTEINS, where FC\textsubscript{mlp} variants outperform baselines by more than 6--10 percentage points, highlighting the ability to capture complex biochemical patterns. Social network datasets also benefit significantly: on IMDB-BINARY, accuracy improves from 50.90\% (GCN) to 69.40\% with FC\textsubscript{mlp}GCN, representing an 18.50-point increase.

These findings confirm that the observed performance gains are not solely due to deeper GNN backbones, but rather by the explicit incorporation of topological features through the FC block. The consistency of improvements across multiple backbones demonstrates the robustness of functional connectivity augmentation in deeper architectures.

\begin{table*}[ht]
\footnotesize
\centering
\begin{tabular}{lcccccc}
\toprule
Model & ENZYMES & BZR & COX2 & MUTAG & PROTEINS & IMDB\_BINARY \\
\midrule
FC\textsubscript{cnn}GCN & 36.47$\pm$2.06 & 83.85$\pm$2.23 & 79.19$\pm$2.27 & \boldmax{84.95}$\pm$5.84 & 74.49$\pm$2.16  & \boldmax{70.20}$\pm$0.93 \\
FC\textsubscript{mlp}GCN & \boldmax{43.27}$\pm$4.47 & \boldmax{84.69}$\pm$2.04 & \boldmax{81.87}$\pm$2.38 & 84.95$\pm$4.18 & \boldmax{76.45}$\pm$2.40 & 69.40$\pm$1.46 \\
GCN & 26.73$\pm$3.02 & 82.52$\pm$3.25 & 81.19$\pm$3.72 & 78.32$\pm$3.82 & 70.28$\pm$2.51 & 50.90$\pm$1.39 \\
\midrule
FC\textsubscript{cnn}GIN & 44.60$\pm$5.99 & 83.16$\pm$3.45 & 80.30$\pm$4.25 & \boldmax{86.21}$\pm$3.00 & 75.05$\pm$2.21 & 70.00$\pm$2.30\\
FC\textsubscript{mlp}GIN & \boldmax{48.03}$\pm$4.99 & \boldmax{84.35}$\pm$1.84 & 81.49$\pm$3.85 & 84.84$\pm$4.72 & \boldmax{75.70}$\pm$2.02 & 70.70$\pm$1.69\\
GIN & 45.23$\pm$3.38 & 83.90$\pm$2.48 & \boldmax{82.43}$\pm$3.75 & 83.16$\pm$4.47 & 71.50$\pm$2.43 & \boldmax{71.20}$\pm$2.38\\
\midrule
FC\textsubscript{cnn}GAT & 34.33$\pm$2.56 & 78.86$\pm$1.45 & \boldmax{77.45}$\pm$0.43 & 85.05$\pm$5.35 & 74.33$\pm$2.14 & \boldmax{69.83}$\pm$0.29\\
FC\textsubscript{mlp}GAT & \boldmax{37.57}$\pm$3.95 & \boldmax{79.11}$\pm$2.44 & 77.11$\pm$1.28 & \boldmax{86.53}$\pm$3.89 & \boldmax{75.00}$\pm$2.46 & 67.62$\pm$1.87\\
GAT & 23.53$\pm$2.56 & 79.01$\pm$1.01 & 76.13$\pm$7.50 & 70.74$\pm$3.75 & 70.06$\pm$1.67 & 59.47$\pm$3.21\\
\midrule
FC\textsubscript{cnn}GSAGE & 40.80$\pm$3.01 & 82.07$\pm$2.80 & 82.34$\pm$3.35 & 85.16$\pm$5.76 & 74.49$\pm$2.41 & \boldmax{68.13}$\pm$2.57\\
FC\textsubscript{mlp}GSAGE & \boldmax{46.20}$\pm$3.92 & \boldmax{83.56}$\pm$2.20 & \boldmax{83.36}$\pm$3.01 & \boldmax{86.21}$\pm$5.63 & \boldmax{76.04}$\pm$2.18 & 67.24$\pm$4.21\\
GSAGE & 36.13$\pm$4.70 & 81.73$\pm$3.56 & 83.15$\pm$3.59 & 81.37$\pm$5.62 & 69.38$\pm$2.49 & 52.33$\pm$1.32\\
\midrule
FC\textsubscript{cnn}ChebNet & 47.50$\pm$2.79 & 82.72$\pm$7.45 & 79.00$\pm$4.24 & 79.23$\pm$4.74 & 75.02$\pm$1.15 & \boldmax{69.80}$\pm$0.50\\
FC\textsubscript{mlp}ChebNet & \boldmax{49.33}$\pm$2.66 & 84.69$\pm$2.54 & \boldmax{81.15}$\pm$4.29 & \boldmax{80.85}$\pm$2.58 & \boldmax{75.65}$\pm$1.73 & 68.70$\pm$1.29\\
ChebNet & 48.17$\pm$0.62 & \boldmax{85.19}$\pm$1.56 & 80.94$\pm$2.29 & 78.21$\pm$1.79 & 69.63$\pm$2.26 & 59.90$\pm$3.09\\
\midrule
FC\textsubscript{cnn}GATv2 & \boldmax{34.00}$\pm$4.84 & 77.78$\pm$3.12 & 78.38$\pm$0.98 & 81.38$\pm$3.81 & \boldmax{75.11}$\pm$1.25 & 66.10$\pm$8.12\\
FC\textsubscript{mlp}GATv2 & 33.67$\pm$2.67 & 77.53$\pm$4.51 & \boldmax{79.23}$\pm$1.32 & \boldmax{82.97}$\pm$3.23 & 74.84$\pm$0.71& \boldmax{70.20}$\pm$2.64\\
GATv2 & 23.50$\pm$5.01 & \boldmax{78.77}$\pm$0.49 & 78.16$\pm$0.41 & 69.77$\pm$7.28 & 71.07$\pm$1.11& 49.40$\pm$1.50\\
\midrule
FC\textsubscript{cnn}ARMA & 40.67$\pm$4.90 & \boldmax{82.47}$\pm$0.92 & 82.44$\pm$2.21 & \boldmax{80.85}$\pm$3.93 & 74.84$\pm$1.48 & \boldmax{70.30}$\pm$2.01\\
FC\textsubscript{mlp}ARMA & \boldmax{46.67}$\pm$4.35 & 81.23$\pm$3.78 & 81.57$\pm$3.13 & 80.33$\pm$4.23 & \boldmax{75.11}$\pm$1.42 & 69.30$\pm$2.06 \\
ARMA & 37.50$\pm$4.71 & 82.22$\pm$2.29 & \boldmax{83.50}$\pm$2.04 & 77.68$\pm$4.52 & 71.07$\pm$1.89 & 53.40$\pm$3.93\\
\midrule
FC\textsubscript{cnn}TAG & 44.17$\pm$3.91 & \boldmax{82.72}$\pm$2.82 & 81.14$\pm$3.36 & \boldmax{81.92}$\pm$3.09 & 75.29$\pm$1.91 & 69.00$\pm$1.58\\
FC\textsubscript{mlp}TAG & \boldmax{48.83}$\pm$3.67 & 82.72$\pm$3.66 & 81.79$\pm$2.42 & 79.80$\pm$3.89 & \boldmax{75.56}$\pm$1.88 & \boldmax{69.40}$\pm$2.13\\
TAG & 42.33$\pm$4.52 & 81.48$\pm$4.20 & \boldmax{82.87}$\pm$2.37 & 75.58$\pm$4.95 & 70.62$\pm$2.23 & 50.30$\pm$1.75\\
\midrule
FC\textsubscript{cnn}GraphConv & 44.67$\pm$3.40 & 82.47$\pm$4.51 & 81.36$\pm$3.37 & \boldmax{85.14}$\pm$5.36 & 75.11$\pm$1.13 & 69.00$\pm$0.89\\
FC\textsubscript{mlp}GraphConv & \boldmax{49.50}$\pm$1.94 & 80.49$\pm$3.86 & \boldmax{83.08}$\pm$1.75 & 84.61$\pm$2.93 & \boldmax{75.30}$\pm$2.20 & \boldmax{70.20}$\pm$4.53\\
GraphConv & 38.00$\pm$5.74 & \boldmax{83.46}$\pm$3.54 & 82.01$\pm$1.33 & 82.46$\pm$4.85 & 68.60$\pm$1.21 & 67.60$\pm$3.60\\
\midrule
FC\textsubscript{cnn} & 28.63$\pm$3.01 & 79.06$\pm$0.24 & 77.66$\pm$1.01 & 82.00$\pm$4.92 & 73.69$\pm$2.44 & 70.69$\pm$1.27\\
FC\textsubscript{mlp} & 28.57$\pm$3.30 & 79.01$\pm$1.01 & 77.62$\pm$0.21 & 82.16$\pm$3.89 & 74.33$\pm$2.66 & 68.69$\pm$2.13\\
\bottomrule
\end{tabular}
\caption{Comparison of model performance across 6 datasets (ENZYMES, BZR, COX2, MUTAG, PROTEINS, IMDB\_BINARY), reported as average accuracy scores$\pm$standard deviation. Use GNN 3 layers as baselines}
\label{tab:3layer-6dataset-acc}
\end{table*}

\begin{table*}[ht]
\footnotesize
\centering
\begin{tabular}{lcccccc}
\toprule
Model & ENZYMES & BZR & COX2 & MUTAG & PROTEINS & IMDB\_BINARY \\
\midrule
FC\textsubscript{cnn}GCN & 36.11$\pm$2.22 & 81.92$\pm$2.98 & 75.71$\pm$2.73 & \boldmax{84.72}$\pm$5.78 & 73.62$\pm$2.32 & \boldmax{70.03}$\pm$1.02\\
FC\textsubscript{mlp}GCN & \boldmax{42.36}$\pm$4.70 & \boldmax{82.45}$\pm$3.28 & \boldmax{80.26}$\pm$3.32 & 84.43$\pm$4.44 & \boldmax{76.08}$\pm$2.45 & 68.46$\pm$1.84\\
GCN & 24.36$\pm$2.78 & 80.95$\pm$3.26 & 78.60$\pm$5.24 & 76.19$\pm$5.72 & 68.81$\pm$3.10 & 44.26$\pm$4.91\\
\midrule
FC\textsubscript{cnn}GIN & 44.20$\pm$6.34 & 81.48$\pm$3.89 & 77.99$\pm$4.89 & \boldmax{86.02}$\pm$3.14 & 74.34$\pm$2.38 & 69.79$\pm$2.34\\
FC\textsubscript{mlp}GIN & \boldmax{47.12}$\pm$5.24 & 81.83$\pm$3.04 & 79.08$\pm$5.05 & 84.66$\pm$4.68 & \boldmax{75.26}$\pm$2.19 & 70.36$\pm$1.79\\
GIN & 44.98$\pm$3.44 & \boldmax{82.19}$\pm$2.75 & \boldmax{80.74}$\pm$4.31 & 82.52$\pm$4.68 & 70.58$\pm$2.59 & \boldmax{70.79}$\pm$2.39\\
\midrule
FC\textsubscript{cnn}GAT & 33.38$\pm$3.40 & 70.59$\pm$1.69 & 67.79$\pm$0.21 & 84.71$\pm$5.36 & 73.34$\pm$2.39 & 69.23$\pm$4.24\\
FC\textsubscript{mlp}GAT & \boldmax{36.60}$\pm$3.99 & \boldmax{72.99}$\pm$3.36 & \boldmax{68.32}$\pm$1.62 & \boldmax{86.21}$\pm$4.16 & \boldmax{74.39}$\pm$2.70 & \boldmax{76.82}$\pm$1.23\\
GAT & 19.61$\pm$2.07 & 69.75$\pm$1.01 & 66.92$\pm$4.76 & 62.62$\pm$7.04 & 68.92$\pm$2.17 & 62.34$\pm$1.29\\
\midrule
FC\textsubscript{cnn}GSAGE & 40.00$\pm$3.13 & 79.71$\pm$3.92 & 80.34$\pm$4.08 & 84.99$\pm$5.63 & 73.64$\pm$2.58 & \boldmax{69.23}$\pm$1.20 \\
FC\textsubscript{mlp}GSAGE & \boldmax{45.28}$\pm$4.19 & \boldmax{80.82}$\pm$3.15 & \boldmax{81.59}$\pm$3.54 & \boldmax{85.78}$\pm$5.83 & \boldmax{75.70}$\pm$2.18 & 68.98$\pm$1.87\\
GSAGE & 34.15$\pm$5.32 & 79.66$\pm$4.04 & 81.28$\pm$4.43 & 80.19$\pm$6.40 & 68.42$\pm$3.16 & 63.97$\pm$1.98\\
\midrule
FC\textsubscript{cnn}ChebNet & 46.82$\pm$2.68  & 82.26$\pm$6.66 & 78.30$\pm$3.66 & 79.06$\pm$4.66 & 74.52$\pm$1.37 & \boldmax{69.46}$\pm$0.72\\
FC\textsubscript{mlp}ChebNet & \boldmax{48.79}$\pm$3.06 & 83.73$\pm$3.09 & \boldmax{80.02}$\pm$4.50 & \boldmax{80.51}$\pm$2.88  & \boldmax{75.44}$\pm$1.84 & 67.88$\pm$1.77\\
ChebNet & 47.85$\pm$0.91 & \boldmax{83.85}$\pm$2.41 & 78.47$\pm$2.98 & 78.22$\pm$1.97 & 68.82$\pm$2.21 & 59.56$\pm$2.72\\
\midrule
FC\textsubscript{cnn}GATv2 & \boldmax{34.08}$\pm$4.69 & 70.11$\pm$3.09 & 70.09$\pm$2.45 & 81.01$\pm$3.91 & \boldmax{74.71}$\pm$1.31 & 62.40$\pm$14.57\\
FC\textsubscript{mlp}GATv2 & 32.74$\pm$2.34 & \boldmax{71.85}$\pm$4.46 & \boldmax{72.35}$\pm$2.94 & \boldmax{82.70}$\pm$3.38 & 74.41$\pm$0.71 & \boldmax{69.86}$\pm$2.81\\
GATv2 & 18.39$\pm$5.09 & 69.41$\pm$0.68 & 68.58$\pm$0.56 & 61.71$\pm$11.23 & 69.78$\pm$2.08 & 39.76$\pm$3.39\\
\midrule
FC\textsubscript{cnn}ARMA & 39.97$\pm$5.06 & 79.20$\pm$1.16 & 80.53$\pm$2.30 & \boldmax{80.54}$\pm$4.09 & 74.42$\pm$1.75 & \boldmax{69.84}$\pm$2.31\\
FC\textsubscript{mlp}ARMA & \boldmax{46.08}$\pm$4.21 & 79.66$\pm$3.24 & 80.83$\pm$2.87 & 79.99$\pm$4.08 & \boldmax{74.93}$\pm$1.45 & 68.89$\pm$1.99\\
ARMA & 35.85$\pm$5.24 & \boldmax{80.62}$\pm$2.28 & \boldmax{82.09}$\pm$2.27 & 77.42$\pm$4.72 & 69.74$\pm$2.40 & 46.08$\pm$8.99\\
\midrule
FC\textsubscript{cnn}TAG & 43.69$\pm$4.09 & 81.14$\pm$3.35 & 79.71$\pm$3.75 & \boldmax{81.58}$\pm$3.40 & 74.88$\pm$2.10 & 68.23$\pm$2.01\\
FC\textsubscript{mlp}TAG & \boldmax{48.31}$\pm$3.09 & 81.33$\pm$3.26 & 79.03$\pm$2.58 & 79.73$\pm$3.66 & \boldmax{75.37}$\pm$1.94 & \boldmax{68.85}$\pm$2.02\\
TAG & 41.36$\pm$5.11 & \boldmax{81.52}$\pm$3.72 & \boldmax{81.17}$\pm$2.26 & 74.65$\pm$5.32 & 69.84$\pm$2.19 & 42.34$\pm$3.89\\
\midrule
FC\textsubscript{cnn}GraphConv & 44.09$\pm$4.14 & 81.41$\pm$4.64 & 80.09$\pm$3.58 & \boldmax{85.20}$\pm$5.32 & 74.63$\pm$1.19 & 68.50$\pm$1.02\\
FC\textsubscript{mlp}GraphConv & \boldmax{48.92}$\pm$2.03 & 79.57$\pm$3.61 & \boldmax{82.19}$\pm$1.35 & 84.60$\pm$2.97 & \boldmax{75.15}$\pm$2.21 & \boldmax{69.87}$\pm$4.57\\
GraphConv & 37.40$\pm$6.29 & \boldmax{82.92}$\pm$3.46 & 80.85$\pm$0.85 & 82.34$\pm$5.09 & 63.25$\pm$2.19 & 67.52$\pm$3.51\\
\midrule
FC\textsubscript{cnn} & 27.52$\pm$3.08 & 71.10$\pm$1.03 & 67.89$\pm$1.01 & 81.73$\pm$5.07 & 72.86$\pm$2.88 & 68.22$\pm$0.31\\
FC\textsubscript{mlp} & 27.32$\pm$3.05 & 70.75$\pm$1.01 & 67.87$\pm$0.10 & 80.81$\pm$4.22 & 73.68$\pm$2.92 & 67.41$\pm$1.29\\
\bottomrule
\end{tabular}
\caption{Comparison of model performance across 6 datasets (ENZYMES, BZR, COX2, MUTAG, PROTEINS, IMDB\_BINARY), reported as average weighted F1 scores$\pm$standard deviation. Use GNN 3 layers as baselines}
\label{tab:3layer-6dataset-f1}
\end{table*}

\section{SOTA baseline Comparisons}
\subsection{Additional Results for SOTA comparisons}
Table~\ref{tab:sota_core_acc} and Table~\ref{tab:sota_core_f1} provide the complete results corresponding to the comparisons reported in Table~3 of the main text. We report both average accuracy and average weighted F1 score (mean $\pm$ standard deviation across 100 runs). These results confirm the robustness of FC-GNN performance across different evaluation metrics.

\begin{table*}[t]
\footnotesize
\centering
\begin{tabular}{lccccc}
\toprule
Model & COLLAB & IMDB\_BINARY & MUTAG & ENZYMES & PROTEINS \\
\midrule
GCN (None) & 57.88$\pm$1.26 & 47.94$\pm$4.29 & 74.25$\pm$9.60 & 26.45$\pm$5.21 & 69.62$\pm$4.47 \\
GCN + DIGL & 53.69$\pm$2.09 & 56.37$\pm$7.71 & 71.20$\pm$9.38 & 25.07$\pm$5.32 & 69.76$\pm$4.12 \\
GCN + SDRF & 64.13$\pm$2.28 & 48.58$\pm$4.57 & 71.85$\pm$9.50 & 24.35$\pm$5.72 & 70.12$\pm$4.16 \\
GCN + FoSR & 64.16$\pm$2.16 & 48.20$\pm$4.27 & 75.80$\pm$8.36 & 20.75$\pm$5.01 & 70.80$\pm$4.21 \\
GCN + PANDA & 66.28$\pm$3.10 & 61.96$\pm$5.68 & \boldmax{82.35$\pm$8.01} & 27.77$\pm$5.51 & \boldmax{74.79$\pm$4.26} \\
\midrule
FC\textsubscript{mlp}GCN & \boldmax{72.80$\pm$1.97} & \boldmax{69.84$\pm$4.69} & 79.70$\pm$8.88 & \boldmax{37.78$\pm$5.44} & 73.25$\pm$3.82 \\
\midrule
GIN (None) & 69.50$\pm$2.64 & 69.62$\pm$4.16 & 82.40$\pm$8.65 & 33.90$\pm$5.48 & 70.09$\pm$4.67 \\
GIN + DIGL & 54.28$\pm$2.45 & 52.42$\pm$8.53 & 71.55$\pm$11.48 & 28.48$\pm$5.46 & 66.45$\pm$5.73 \\
GIN + SDRF & 63.29$\pm$8.11 & 52.77$\pm$8.66 & 73.45$\pm$11.76 & 29.48$\pm$5.67 & 67.79$\pm$5.59 \\
GIN + FoSR & 63.50$\pm$7.84 & 50.57$\pm$6.90 & 76.85$\pm$12.56 & 22.00$\pm$5.28 & 70.29$\pm$7.06 \\
GIN + PANDA & 69.93$\pm$2.59 & 66.54$\pm$4.97 & \boldmax{83.85$\pm$7.90} & 34.52$\pm$5.68 & \boldmax{73.93$\pm$3.66} \\
\midrule
FC\textsubscript{mlp}GIN & \boldmax{74.50$\pm$1.90} & \boldmax{70.39$\pm$4.89} & 82.65$\pm$8.53 & \boldmax{43.55$\pm$6.33} & 73.03$\pm$3.73 \\
\bottomrule
\end{tabular}
\caption{Comparison of FC\textsubscript{mlp} with SOTA models under 80\%/10\%/10\% data split (Training/Validation/Testing) across 5 datasets (COLLAB, IMDB\_BINARY, MUTAG, ENZYMES, PROTEINS), reported as average accuracy scores$\pm$standard deviation.}
\label{tab:sota_core_acc}
\end{table*}

\begin{table*}[t]
\footnotesize
\centering
\begin{tabular}{lccccc}
\toprule
Model & COLLAB & IMDB\_BINARY & MUTAG & ENZYMES & PROTEINS \\
\midrule
GCN (None) & 59.32$\pm$2.35 & 35.76$\pm$7.31 & 73.09$\pm$10.24 & 23.32$\pm$4.86 & 68.73$\pm$4.71 \\
GCN + DIGL & 39.98$\pm$2.38 & 51.63$\pm$13.51 & 69.54$\pm$10.19 & 22.00$\pm$5.18 & 68.39$\pm$4.50 \\
GCN + SDRF & 62.23$\pm$2.47 & 34.19$\pm$8.26 & 70.17$\pm$10.37 & 21.20$\pm$5.55 & 68.79$\pm$4.55 \\
GCN + FoSR & 62.32$\pm$2.35 & 32.96$\pm$7.29 & 74.20$\pm$9.42 & 17.64$\pm$4.80 & 69.50$\pm$4.64 \\
GCN + PANDA & 64.92$\pm$3.42 & 61.63$\pm$5.83 & \boldmax{81.80$\pm$8.31} & 26.86$\pm$5.66 & \boldmax{74.27$\pm$4.43} \\
\midrule
FC\textsubscript{mlp}GCN & 
\boldmax{72.73$\pm$1.99} & \boldmax{69.38$\pm$4.88} & 79.47$\pm$8.83 & \boldmax{37.10$\pm$5.75} & 72.69$\pm$3.99 \\
\midrule
GIN (None) & 71.30$\pm$4.21 & 69.51$\pm$4.22 & 82.28$\pm$8.62 & 32.89$\pm$5.72 & 69.37$\pm$4.89 \\
GIN + DIGL & 42.16$\pm$2.86 & 41.31$\pm$15.07 & 68.20$\pm$14.92 & 26.90$\pm$5.55 & 62.48$\pm$9.87 \\
GIN + SDRF & 58.09$\pm$14.98 & 40.52$\pm$15.57 & 70.91$\pm$14.78 & 27.52$\pm$5.87 & 64.76$\pm$9.19 \\
GIN + FoSR & 58.30$\pm$14.71 & 36.13$\pm$12.00 & 74.42$\pm$15.81 & 19.81$\pm$5.79 & 66.96$\pm$11.36 \\
GIN + PANDA & 69.24$\pm$2.67 & 66.24$\pm$5.06 & \boldmax{83.48$\pm$8.60} & 34.01$\pm$6.00 & \boldmax{73.43$\pm$3.97} \\
\midrule
FC\textsubscript{mlp}GIN & \boldmax{74.26$\pm$1.98} & \boldmax{70.00$\pm$5.07} & 82.44$\pm$8.82 & \boldmax{42.73$\pm$6.64} & 72.68$\pm$3.83 \\
\bottomrule
\end{tabular}
\caption{Comparison of FC\textsubscript{mlp} model performance with SOTA models under 80\%/10\%/10\% data split (Training/Validation/Testing) across 5 datasets (COLLAB, IMDB\_BINARY, MUTAG, ENZYMES, PROTEINS), reported as average weighted F1 score$\pm$standard deviation.}
\label{tab:sota_core_f1}
\end{table*}

\subsection{SOTA few-shot experiments}
We evaluated our methods under extreme data scarcity conditions using a few-shot learning setup to assess model robustness when labeled data is severely limited—a common scenario in real-world applications where graph annotation is expensive or domain expertise is required. This challenging setting tests whether architectural innovations provide genuine inductive bias or merely fit abundant training data. All experimental settings remain identical to our main SOTA experiments except for three necessary adjustments: (1) the train/val/test split ratio, (2) dataset-specific epoch numbers, and (3) dataset-specific batch sizes.

\paragraph{Train/Val/Test Split Ratio.} The SOTA few-shot experiment uses a 5\%/5\%/90\% train/val/test split. This extreme reduction creates a challenging scenario where models must learn from minimal examples.

\paragraph{Epochs by Dataset.} We adapt epochs by dataset complexity and size. This adjustment ensures each model sees the limited training data sufficient times to learn meaningful patterns, with more complex datasets requiring additional iterations. Table~\ref{tab:sota_extra_epoch} summarizes the epoch settings for each dataset along with the corresponding rationale.

\begin{table*}[ht]
\centering
\setlength{\tabcolsep}{1mm}
\begin{tabular}{lcl}
\hline
\textbf{Dataset} & \textbf{Epochs} & \textbf{Rationale} \\
\hline
MUTAG & 300 & \makecell[l]{Small size \\ Simple patterns} \\
\hline
ENZYMES & 500 & \makecell[l]{Multi-class (6 classes) \\ High Complexity} \\
\hline
PROTEINS & 400 & \makecell[l]{Binary classification \\ Moderate complexity} \\
\hline
IMDB-BINARY & 600 & \makecell[l]{No node features \\ Hard to learn} \\
\hline
COLLAB & 800 & \makecell[l]{Dense graphs ($\sim$2,457 edges) \\ Higher complexity} \\
\hline
\end{tabular}
\caption{Epoch configuration for selected datasets in SOTA few-shot experiments.}
\label{tab:sota_extra_epoch}
\end{table*}

\paragraph{Batch Size by Dataset.} Batch sizes are constrained by the small training set sizes to ensure valid gradient computation. These modifications follow standard few-shot learning practices where longer training and smaller batches help models extract maximum information from limited samples while avoiding degenerate cases where batch size exceeds training set size. Table~\ref{tab:sota_extra_batch} summarizes the batch size for each dataset.

\begin{table*}[ht]
\centering
\setlength{\tabcolsep}{1mm}
\begin{tabular}{lccc}
\hline
\textbf{Dataset} & \textbf{Total Graphs} & \textbf{5\% Train} & \textbf{Batch Size} \\
\hline
MUTAG & 188 & ~9 & \textbf{8} \\
ENZYMES & 600 & ~30 & \textbf{16} \\
PROTEINS & 1,113 & ~56 & \textbf{32} \\
IMDB-BINARY & 1,000 & ~50 & \textbf{32} \\
COLLAB & 5,000 & ~250 & \textbf{64} \\
\hline
\end{tabular}
\caption{Batch size configuration for selected datasets in SOTA few-shot experiments.}
\label{tab:sota_extra_batch}
\end{table*}

Table~\ref{tab:sota_extra_acc} and Table~\ref{tab:sota_extra_f1} provide the complete results corresponding to the comparisons reported in Table~4 of the main text. We report both average accuracy and average weighted F1 score (mean $\pm$ standard deviation across 100 runs).

\begin{table*}[t]
\footnotesize
\centering
\begin{tabular}{lccccc}
\toprule
Model & COLLAB & IMDB\_BINARY & MUTAG & ENZYMES & PROTEINS \\
\midrule
GCN (None) & 52.80$\pm$0.83 & 50.00$\pm$0.93 & 64.70$\pm$8.04 & 20.52$\pm$2.80 & 65.51$\pm$3.12 \\
GCN + DIGL & 52.03$\pm$0.24 & 49.93$\pm$1.42 & 64.88$\pm$7.92 & 19.83$\pm$2.63 & 63.74$\pm$2.90 \\
GCN + SDRF & 61.34$\pm$2.09 & 49.65$\pm$0.44 & 65.28$\pm$7.45 & 19.32$\pm$2.49 & 63.86$\pm$3.40 \\
GCN + FoSR & 57.31$\pm$4.23 & 49.65$\pm$0.44 & 66.12$\pm$7.38 & 18.85$\pm$2.47 & 64.55$\pm$3.11 \\
GCN + PANDA & 63.33$\pm$2.18 & 56.89$\pm$3.60 & \boldmax{75.61$\pm$6.11}& 21.38$\pm$2.33 & \boldmax{70.53$\pm$2.28} \\
\midrule
FC\textsubscript{mlp}GCN & \boldmax{65.32$\pm$1.79} & \boldmax{62.58$\pm$3.57} & 71.84$\pm$7.84 & \boldmax{23.60$\pm$2.48} & 68.10$\pm$3.55 \\
\midrule
GIN (None) & \boldmax{67.04$\pm$1.23} & \boldmax{65.53$\pm$4.86} & 67.64$\pm$6.91 & 22.89$\pm$3.06 & 62.67$\pm$3.39 \\
GIN + DIGL & 53.81$\pm$1.89 & 54.23$\pm$6.21 & 62.95$\pm$11.57 & 22.00$\pm$2.76 & 63.23$\pm$4.25 \\
GIN + SDRF & 56.36$\pm$5.86 & 52.60$\pm$5.97 & 63.29$\pm$12.02 & 21.79$\pm$2.72 & 64.00$\pm$3.97 \\
GIN + FoSR & 57.34$\pm$5.89 & 51.85$\pm$5.10 & 66.82$\pm$11.02 & 20.34$\pm$1.99 & 67.14$\pm$4.20 \\
GIN + PANDA & 65.21$\pm$1.98 & 62.39$\pm$5.42 & \boldmax{73.81$\pm$6.84} & 23.00$\pm$2.76 & \boldmax{68.20$\pm$3.37} \\
\midrule
FC\textsubscript{mlp}GIN & 66.80$\pm$1.53 & 62.97$\pm$3.58 & 70.56$\pm$8.00 & \boldmax{25.27$\pm$3.03} & 66.99$\pm$3.53 \\
\bottomrule
\end{tabular}
\caption{Comparison of FC\textsubscript{mlp} with SOTA models under 5\%/5\%/90\% data split (Training/Validation/Testing), reported as average accuracy scores$\pm$standard deviation.}
\label{tab:sota_extra_acc}
\end{table*}

\begin{table*}[t]
\footnotesize
\centering
\begin{tabular}{lccccc}
\toprule
Model & COLLAB & IMDB\_BINARY & MUTAG & ENZYMES & PROTEINS \\
\midrule
GCN (None) & 41.85$\pm$2.50 & 37.33$\pm$3.89 & 61.27$\pm$8.61 & 17.09$\pm$3.17 & 63.90$\pm$4.00 \\
GCN + DIGL & 35.67$\pm$0.42 & 33.79$\pm$3.91 & 54.59$\pm$10.77 & 14.59$\pm$3.53 & 60.16$\pm$4.68 \\
GCN + SDRF & 59.12$\pm$4.81 & 32.94$\pm$0.48 & 55.53$\pm$10.32 & 13.90$\pm$3.48 & 60.09$\pm$5.46 \\
GCN + FoSR & 47.62$\pm$9.45 & 32.94$\pm$0.48 & 56.88$\pm$10.65 & 13.16$\pm$3.33 & 60.97$\pm$5.17 \\
GCN + PANDA & 62.27$\pm$2.88 & 54.67$\pm$6.05 & \boldmax{73.51$\pm$8.31} & 19.30$\pm$2.59 & \boldmax{69.59$\pm$2.69} \\
\midrule
FC\textsubscript{mlp}GCN & \boldmax{65.14$\pm$1.86} & \boldmax{61.94$\pm$3.89} & 70.35$\pm$9.31 & \boldmax{22.03$\pm$2.57} & 67.80$\pm$3.55 \\
\midrule
GIN (None) & 66.12$\pm$2.04 & \boldmax{64.24$\pm$7.31} & 64.95$\pm$7.93 & 21.58$\pm$3.13 & 61.96$\pm$3.35 \\
GIN + DIGL & 41.81$\pm$1.66 & 44.94$\pm$12.73 & 54.53$\pm$15.15 & 19.60$\pm$3.20 & 57.96$\pm$8.55 \\
GIN + SDRF & 45.61$\pm$12.78 & 38.92$\pm$12.17 & 55.61$\pm$15.74 & 19.64$\pm$3.03 & 60.01$\pm$7.56 \\
GIN + FoSR & 48.12$\pm$12.81 & 37.50$\pm$10.59 & 61.28$\pm$14.82 & 17.91$\pm$2.25 & 63.29$\pm$8.01 \\
GIN + PANDA & 64.33$\pm$1.97 & 60.21$\pm$8.80 & \boldmax{71.41$\pm$9.10} & 21.57$\pm$2.87 & \boldmax{67.61$\pm$3.62} \\
\midrule
FC\textsubscript{mlp}GIN & \boldmax{66.46$\pm$1.73} & 62.39$\pm$3.66 & 69.14$\pm$9.25 & \boldmax{23.76$\pm$3.09} & 66.66$\pm$3.49 \\
\bottomrule
\end{tabular}
\caption{Comparison of FC\textsubscript{mlp} with SOTA models under 5\%/5\%/90\% data split (Training/Validation/Testing), reported as average weighted F1 scores$\pm$standard deviation.}
\label{tab:sota_extra_f1}
\end{table*}

\section{Runtime and Memory Analysis}

We provide scalability arguments in the main text. Here we present detailed computational cost comparisons, reporting runtime and GPU memory usage across 10 benchmark datasets.

\subsection{Runtime Analysis}
\Cref{tab:runtime_2layer,tab:runtime_3layer} reveal that FC\textsubscript{mlp} variants incur only modest computational overhead while delivering substantial performance improvements. Standalone FC-GNN models (FC\textsubscript{cnn} and FC\textsubscript{mlp}) demonstrate the highest efficiency, with FC\textsubscript{mlp} achieving the fastest runtimes across most datasets—notably 4.98 seconds on ENZYMES and 8.63 seconds on PROTEINS\_full in the 2-layer setting, and 6.43 seconds and 11.28 seconds respectively in the 3-layer setting. Among integrated architectures, FC\textsubscript{mlp}GCN and FC\textsubscript{mlp}GIN show only moderate runtime increases relative to their baselines: for example, on PROTEINS with 2 layers, FC\textsubscript{mlp}GIN requires 23.29 seconds versus GIN’s 18.63 seconds (a 24.69\% increase), while on PROTEINS\_full with 3 layers, FC\textsubscript{mlp}GIN runs in 19.74 seconds compared to GIN’s 20.58 seconds. The relative overhead is consistent across depths, typically 10–20\% for FC\textsubscript{mlp}GCN. The largest slowdowns occur for complex attention-based architectures (e.g., GAT) on large datasets (NCI1, NCI109), where runtime increases are compounded when using attention mechanisms. For instance, on the NCI1 dataset, the runtime for the 2-layer FC\textsubscript{mlp}GAT is 96.02 seconds compared to 88.41 seconds for GAT, and the runtime for the 3-layer FC\textsubscript{mlp}GAT is 149.08 seconds versus 133.87 seconds for GAT. Importantly, for smaller molecular datasets such as MUTAG and BZR, runtime differences remain negligible (generally under 2–3 seconds). These results confirm that functional connectivity augmentation scales efficiently with model depth and dataset size, maintaining practical runtimes suitable for real-world deployment.

\begin{table*}[ht]
\footnotesize
\centering
\begin{tabular}{lccccc}
\toprule
Model & DD & ENZYMES & BZR & COX2 & MUTAG \\
\midrule
FC\textsubscript{cnn}GCN & 30.93$\pm$0.90 & 13.00$\pm$1.31 & 10.48$\pm$0.61 & 10.71$\pm$0.55 & 5.16$\pm$0.10 \\
FC\textsubscript{mlp}GCN & 34.27$\pm$0.59 & 11.66$\pm$0.83 & 8.36$\pm$0.48 & 8.95$\pm$0.64 & 4.01$\pm$0.13 \\
GCN & 29.62$\pm$1.74 & 10.99$\pm$0.65 & 7.23$\pm$0.05 & 7.55$\pm$0.06 & 3.53$\pm$0.06 \\
\midrule
FC\textsubscript{cnn}GIN & 29.78$\pm$0.33 & 13.28$\pm$1.52 & 10.50$\pm$0.33 & 10.92$\pm$0.35 & 5.23$\pm$0.11 \\
FC\textsubscript{mlp}GIN & 29.46$\pm$1.17 & 11.10$\pm$0.99 & 8.05$\pm$0.08 & 8.28$\pm$0.03 & 4.21$\pm$0.08 \\
GIN & 29.09$\pm$0.46 & 9.90$\pm$0.63 & 7.48$\pm$0.31 & 7.57$\pm$0.27 & 3.46$\pm$0.05 \\
\midrule
FC\textsubscript{cnn}GAT & 34.24$\pm$0.96 & 17.75$\pm$2.36 & 16.14$\pm$0.11 & 16.52$\pm$0.15 & 7.69$\pm$0.13 \\
FC\textsubscript{mlp}GAT & 34.21$\pm$2.06 & 16.68$\pm$1.80 & 12.69$\pm$0.13 & 12.71$\pm$0.17 & 6.40$\pm$0.05 \\
GAT & 32.95$\pm$0.89 & 14.78$\pm$1.62 & 11.58$\pm$0.28 & 11.70$\pm$0.24 & 5.53$\pm$0.07 \\
\midrule
FC\textsubscript{cnn}GraphSAGE & 29.74$\pm$0.28 & 13.16$\pm$1.66 & 9.83$\pm$0.06 & 10.13$\pm$0.03 & 5.05$\pm$0.06 \\
FC\textsubscript{mlp}GraphSAGE & 31.01$\pm$0.83 & 11.17$\pm$1.00 & 8.82$\pm$0.31 & 9.06$\pm$0.28 & 3.78$\pm$0.04 \\
GraphSAGE & 27.70$\pm$0.30 & 10.16$\pm$0.76 & 7.76$\pm$0.20 & 8.07$\pm$0.17 & 3.54$\pm$0.05 \\
\midrule
FC\textsubscript{cnn} & 20.31$\pm$0.33 & 7.80$\pm$2.57 & 6.45$\pm$0.59 & 7.08$\pm$1.01 & 3.41$\pm$0.03 \\
FC\textsubscript{mlp} & 18.21$\pm$0.27 & 4.98$\pm$0.52 & 4.92$\pm$0.29 & 5.60$\pm$0.64 & 2.79$\pm$0.09 \\
\bottomrule
\toprule
Model & NCI1 & NCI109 & PROTEINS & PROTEINS\_full & PTC\_MR \\
\midrule
FC\textsubscript{cnn}GCN & 80.90$\pm$13.09 & 80.12$\pm$7.82 & 23.53$\pm$7.23 & 21.44$\pm$1.61 & 9.40$\pm$0.71 \\
FC\textsubscript{mlp}GCN & 60.26$\pm$8.30 & 64.07$\pm$6.88 & 18.29$\pm$1.94 & 20.07$\pm$1.60 & 6.79$\pm$0.08 \\
GCN & 51.96$\pm$3.92 & 51.76$\pm$4.23 & 17.62$\pm$3.57 & 17.86$\pm$0.76 & 7.00$\pm$0.22 \\
\midrule
FC\textsubscript{cnn}GIN & 70.69$\pm$16.89 & 79.52$\pm$17.59 & 23.74$\pm$6.81 & 21.75$\pm$2.07 & 8.60$\pm$0.11 \\
FC\textsubscript{mlp}GIN & 70.38$\pm$11.65 & 70.24$\pm$12.72 & 23.29$\pm$4.63 & 17.74$\pm$1.20 & 6.73$\pm$0.07 \\
GIN & 53.62$\pm$10.65 & 56.44$\pm$7.37 & 18.63$\pm$4.05 & 16.98$\pm$1.27 & 6.71$\pm$0.19 \\
\midrule
FC\textsubscript{cnn}GAT & 102.17$\pm$21.68 & 94.64$\pm$26.39 & 29.76$\pm$9.84 & 30.12$\pm$3.44 & 13.26$\pm$0.19 \\
FC\textsubscript{mlp}GAT & 96.02$\pm$21.73 & 107.90$\pm$28.74 & 33.88$\pm$9.30 & 24.70$\pm$2.11 & 10.66$\pm$0.29 \\
GAT & 88.41$\pm$20.12 & 87.12$\pm$17.92 & 29.26$\pm$6.01 & 23.92$\pm$2.49 & 9.39$\pm$0.35 \\
\midrule
FC\textsubscript{cnn}GraphSAGE & 77.18$\pm$16.50 & 78.43$\pm$15.06 & 26.05$\pm$7.87 & 22.04$\pm$2.02 & 8.23$\pm$0.11 \\
FC\textsubscript{mlp}GraphSAGE & 62.19$\pm$9.39 & 62.40$\pm$7.80 & 20.81$\pm$3.38 & 18.04$\pm$1.33 & 6.43$\pm$0.07 \\
GraphSAGE & 50.76$\pm$4.98 & 53.13$\pm$5.75 & 16.59$\pm$3.97 & 16.32$\pm$0.90 & 7.17$\pm$0.65 \\
\midrule
FC\textsubscript{cnn} & 53.70$\pm$8.45 & 51.86$\pm$8.98 & 11.83$\pm$3.05 & 14.29$\pm$3.64 & 6.03$\pm$0.53 \\
FC\textsubscript{mlp} & 36.59$\pm$1.95 & 36.10$\pm$1.70 & 8.87$\pm$0.75 & 8.63$\pm$0.74 & 4.70$\pm$0.23 \\
\bottomrule
\end{tabular}
\caption{Comparison of model runtime across 10 datasets (DD, ENZYMES, BZR, COX2, MUTAG, NCI1, NCI109, PROTEINS, PROTEINS\_{full}, PTC\_MR), reported in second unit. Using 2 GNN layers as baseline}
\label{tab:runtime_2layer}
\end{table*}

\begin{table*}[ht]
\footnotesize
\centering
\begin{tabular}{lccccc}
\toprule
Model & DD & ENZYMES & BZR & COX2 & MUTAG \\
\midrule
FC\textsubscript{cnn}GCN & 55.37$\pm$6.11 & 16.19$\pm$0.71 & 11.06$\pm$0.06 & 14.22$\pm$2.00 & 5.55$\pm$0.04 \\
FC\textsubscript{mlp}GCN & 57.15$\pm$1.06 & 12.86$\pm$0.28 & 9.30$\pm$0.13 & 11.45$\pm$1.35 & 4.59$\pm$0.05 \\
GCN & 46.24$\pm$4.03 & 11.59$\pm$0.29 & 8.27$\pm$0.04 & 9.93$\pm$1.23 & 4.10$\pm$0.04 \\
\midrule
FC\textsubscript{cnn}GIN & 52.75$\pm$3.98 & 16.89$\pm$0.21 & 12.44$\pm$0.13 & 14.86$\pm$1.63 & 6.35$\pm$0.17 \\
FC\textsubscript{mlp}GIN & 46.67$\pm$1.95 & 13.57$\pm$0.17 & 9.95$\pm$0.09 & 10.72$\pm$1.24 & 5.02$\pm$0.07 \\
GIN & 44.96$\pm$3.66 & 11.55$\pm$0.22 & 8.33$\pm$0.09 & 9.45$\pm$1.52 & 4.33$\pm$0.06 \\
\midrule
FC\textsubscript{cnn}GAT & 72.10$\pm$6.65 & 27.67$\pm$0.23 & 20.40$\pm$0.05 & 22.66$\pm$4.22 & 10.24$\pm$0.04 \\
FC\textsubscript{mlp}GAT & 68.88$\pm$3.81 & 23.51$\pm$0.20 & 17.32$\pm$0.16 & 17.10$\pm$2.95 & 8.76$\pm$0.03 \\
GAT & 58.51$\pm$5.46 & 21.95$\pm$0.84 & 15.07$\pm$0.05 & 18.56$\pm$2.89 & 7.87$\pm$0.08 \\
\midrule
FC\textsubscript{cnn}GraphSAGE & 53.22$\pm$3.75 & 16.42$\pm$0.66 & 11.51$\pm$0.06 & 14.24$\pm$2.59 & 5.78$\pm$0.06 \\
FC\textsubscript{mlp}GraphSAGE & 47.62$\pm$4.09 & 13.01$\pm$0.59 & 8.98$\pm$0.09 & 9.91$\pm$1.35 & 4.62$\pm$0.10 \\
GraphSAGE & 42.55$\pm$3.86 & 11.30$\pm$0.50 & 7.75$\pm$0.06 & 11.70$\pm$0.58 & 3.90$\pm$0.05 \\
\midrule
FC\textsubscript{cnn} & 34.55$\pm$1.69 & 8.51$\pm$0.64 & 6.36$\pm$0.60 & 7.11$\pm$0.05 & 3.39$\pm$0.04 \\
FC\textsubscript{mlp} & 29.72$\pm$1.40 & 6.43$\pm$0.14 & 5.13$\pm$0.35 & 5.09$\pm$0.06 & 2.44$\pm$0.03 \\
\bottomrule
\toprule
Model & NCI1 & NCI109 & PROTEINS & PROTEINS\_full & PTC\_MR \\
\midrule
FC\textsubscript{cnn}GCN & 96.24$\pm$0.34 & 103.58$\pm$5.87 & 29.02$\pm$1.63 & 24.15$\pm$2.01 & 9.10$\pm$0.11 \\
FC\textsubscript{mlp}GCN & 81.71$\pm$0.29 & 86.99$\pm$3.57 & 22.06$\pm$0.57 & 21.21$\pm$1.53 & 7.60$\pm$0.08 \\
GCN & 73.46$\pm$0.48 & 78.86$\pm$3.43 & 21.74$\pm$1.08 & 24.61$\pm$1.05 & 6.82$\pm$0.07 \\
\midrule
FC\textsubscript{cnn}GIN & 111.10$\pm$1.82 & 116.35$\pm$8.64 & 33.17$\pm$2.26 & 24.60$\pm$2.71 & 10.27$\pm$0.14 \\
FC\textsubscript{mlp}GIN & 85.29$\pm$0.50 & 97.00$\pm$9.10 & 26.64$\pm$1.33 & 19.74$\pm$1.74 & 8.26$\pm$0.06 \\
GIN & 73.22$\pm$1.12 & 84.13$\pm$7.70 & 23.71$\pm$0.67 & 20.58$\pm$2.16 & 6.98$\pm$0.23 \\
\midrule
FC\textsubscript{cnn}GAT & 182.93$\pm$4.94 & 187.34$\pm$4.71 & 51.77$\pm$1.15 & 34.87$\pm$5.89 & 16.58$\pm$0.07 \\
FC\textsubscript{mlp}GAT & 149.08$\pm$2.00 & 157.26$\pm$6.42 & 45.13$\pm$1.29 & 28.40$\pm$2.72 & 14.40$\pm$0.15 \\
GAT & 133.87$\pm$1.42 & 151.75$\pm$17.84 & 39.17$\pm$1.90 & 32.25$\pm$3.05 & 12.61$\pm$0.04 \\
\midrule
FC\textsubscript{cnn}GraphSAGE & 99.44$\pm$0.93 & 104.33$\pm$6.22 & 31.40$\pm$1.25 & 23.34$\pm$2.15 & 9.60$\pm$0.15 \\
FC\textsubscript{mlp}GraphSAGE & 79.64$\pm$0.87 & 85.59$\pm$5.71 & 23.73$\pm$1.23 & 22.43$\pm$2.01 & 7.49$\pm$0.07 \\
GraphSAGE & 69.91$\pm$0.79 & 74.77$\pm$4.55 & 21.67$\pm$1.00 & 17.96$\pm$1.64 & 6.64$\pm$0.14 \\
\midrule
FC\textsubscript{cnn} & 61.77$\pm$4.48 & 62.51$\pm$1.86 & 15.00$\pm$1.24 & 16.15$\pm$0.09 & 12.43$\pm$0.22 \\
FC\textsubscript{mlp} & 49.77$\pm$2.73 & 46.22$\pm$1.76 & 11.91$\pm$0.61 & 11.28$\pm$0.06 & 10.07$\pm$0.20 \\
\bottomrule
\end{tabular}
\caption{Comparison of model runtime across 10 datasets (DD, ENZYMES, BZR, COX2, MUTAG, NCI1, NCI109, PROTEINS, PROTEINS\_{full}, PTC\_MR), reported in second unit. Using 3 GNN layers as baseline}
\label{tab:runtime_3layer}
\end{table*}

\subsection{GPU Memory Usage}
\Cref{tab:gpu_mem_cost_2layer,tab:gpu_mem_cost_3layer} demonstrate that functional connectivity augmentation requires only modest additional memory while yielding significant performance gains. Standalone FC-GNN models exhibit remarkable memory efficiency, consuming just 17.78–31.82 MB across all datasets in the 2-layer setting and 17.68–31.48 MB in the 3-layer setting, making them the most memory-efficient option overall. Memory requirements scale predictably with dataset size and model depth: 3-layer models typically add only 10–20\% memory relative to their 2-layer counterparts. Even for large datasets such as NCI1, all models remain well within typical GPU memory constraints, underscoring the practicality of functional connectivity augmentation for deployment on standard hardware. Taken together, these results confirm that functional connectivity augmentation scales efficiently with model depth, maintaining practical memory footprints suitable for deployment on standard GPU hardware while delivering substantial performance improvements.

\begin{table*}[ht]
\footnotesize
\centering
\begin{tabular}{lccccc}
\toprule
Model & DD & ENZYMES & BZR & COX2 & MUTAG \\
\midrule
FC\textsubscript{cnn}GCN & 195.67$\pm$67.40 & 31.60$\pm$4.21 & 31.25$\pm$0.15 & 31.44$\pm$2.93 & 22.77$\pm$0.88 \\
FC\textsubscript{mlp}GCN & 253.28$\pm$13.48 & 30.74$\pm$4.54 & 27.60$\pm$3.77 & 28.56$\pm$3.95 & 20.31$\pm$1.39 \\
GCN & 153.99$\pm$52.80 & 29.66$\pm$3.46 & 25.32$\pm$2.96 & 29.85$\pm$0.05 & 18.59$\pm$0.73 \\
\midrule
FC\textsubscript{cnn}GIN & 151.67$\pm$27.18 & 32.11$\pm$1.83 & 26.88$\pm$0.93 & 33.20$\pm$0.04 & 24.51$\pm$1.35 \\
FC\textsubscript{mlp}GIN & 95.15$\pm$8.51 & 26.71$\pm$2.55 & 22.91$\pm$3.05 & 29.51$\pm$0.08 & 20.06$\pm$1.36 \\
GIN & 106.60$\pm$29.99 & 27.26$\pm$0.16 & 21.59$\pm$3.26 & 26.97$\pm$3.55 & 21.55$\pm$1.20 \\
\midrule
FC\textsubscript{cnn}GAT & 415.65$\pm$160.92 & 43.25$\pm$8.37 & 36.03$\pm$6.01 & 34.28$\pm$5.83 & 23.77$\pm$0.69 \\
FC\textsubscript{mlp}GAT & 456.79$\pm$99.77 & 44.28$\pm$7.93 & 32.76$\pm$7.28 & 36.82$\pm$8.19 & 22.70$\pm$0.06 \\
GAT & 294.48$\pm$112.91 & 42.09$\pm$10.02 & 23.07$\pm$0.05 & 23.83$\pm$0.02 & 22.55$\pm$3.30 \\
\midrule
FC\textsubscript{cnn}GraphSAGE & 162.51$\pm$28.22 & 29.81$\pm$1.30 & 27.73$\pm$1.70 & 29.97$\pm$1.29 & 22.65$\pm$0.81 \\
FC\textsubscript{mlp}GraphSAGE & 122.69$\pm$28.55 & 25.51$\pm$2.27 & 21.78$\pm$3.00 & 24.07$\pm$3.42 & 20.08$\pm$1.64 \\
GraphSAGE & 109.89$\pm$22.46 & 24.90$\pm$2.07 & 21.52$\pm$2.78 & 24.57$\pm$3.68 & 18.99$\pm$1.25 \\
\midrule
FC\textsubscript{cnn} & 111.15$\pm$9.63 & 27.37$\pm$0.00 & 24.35$\pm$0.00 & 26.35$\pm$0.00 & 21.72$\pm$0.00 \\
FC\textsubscript{mlp} & 31.82$\pm$1.34 & 17.81$\pm$0.19 & 17.97$\pm$0.04 & 18.23$\pm$0.04 & 17.78$\pm$0.00 \\
\bottomrule
\toprule
Model & NCI1 & NCI109 & PROTEINS & PROTEINS\_full & PTC\_MR \\
\midrule
FC\textsubscript{cnn}GCN & 28.70$\pm$2.21 & 27.65$\pm$2.21 & 36.44$\pm$7.04 & 35.44$\pm$8.30 & 22.04$\pm$0.73 \\
FC\textsubscript{mlp}GCN & 29.09$\pm$0.48 & 27.81$\pm$2.53 & 23.51$\pm$0.44 & 30.12$\pm$7.42 & 19.50$\pm$1.78 \\
GCN & 26.16$\pm$3.00 & 26.17$\pm$2.88 & 29.81$\pm$7.06 & 29.77$\pm$6.92 & 19.94$\pm$2.03 \\
\midrule
FC\textsubscript{cnn}GIN & 31.63$\pm$0.09 & 31.92$\pm$0.05 & 33.06$\pm$3.28 & 32.87$\pm$4.49 & 22.73$\pm$1.11 \\
FC\textsubscript{mlp}GIN & 26.67$\pm$2.64 & 28.27$\pm$0.13 & 25.13$\pm$4.96 & 26.54$\pm$3.79 & 19.77$\pm$1.43 \\
GIN & 23.62$\pm$2.95 & 26.27$\pm$2.56 & 26.32$\pm$5.67 & 24.67$\pm$4.66 & 20.60$\pm$1.59 \\
\midrule
FC\textsubscript{cnn}GAT & 34.91$\pm$6.15 & 42.66$\pm$0.25 & 61.33$\pm$17.15 & 63.18$\pm$13.91 & 25.63$\pm$2.88 \\
FC\textsubscript{mlp}GAT & 35.86$\pm$6.02 & 35.98$\pm$6.11 & 52.54$\pm$16.38 & 53.05$\pm$16.73 & 25.24$\pm$3.83 \\
GAT & 37.56$\pm$4.71 & 40.29$\pm$0.23 & 62.83$\pm$16.11 & 57.70$\pm$17.16 & 27.68$\pm$0.19 \\
\midrule
FC\textsubscript{cnn}GraphSAGE & 28.75$\pm$1.00 & 28.51$\pm$1.41 & 30.26$\pm$1.07 & 31.27$\pm$3.35 & 23.56$\pm$0.04 \\
FC\textsubscript{mlp}GraphSAGE & 24.46$\pm$3.08 & 27.01$\pm$0.09 & 23.23$\pm$1.62 & 23.07$\pm$1.61 & 20.59$\pm$1.50 \\
GraphSAGE & 23.95$\pm$2.82 & 21.07$\pm$2.71 & 22.95$\pm$4.28 & 23.00$\pm$4.57 & 19.38$\pm$1.50 \\
\midrule
FC\textsubscript{cnn} & 25.14$\pm$0.48 & 25.06$\pm$0.17 & 28.10$\pm$0.33 & 28.10$\pm$0.33 & 21.40$\pm$0.23 \\
FC\textsubscript{mlp} & 17.99$\pm$0.13 & 18.08$\pm$0.13 & 17.86$\pm$0.06 & 17.92$\pm$0.16 & 17.68$\pm$0.16 \\
\bottomrule
\end{tabular}
\caption{Comparison of model GPU memory usage across 10 datasets (DD, ENZYMES, BZR, COX2, MUTAG, NCI1, NCI109, PROTEINS, PROTEINS\_{full}, PTC\_MR), reported in MB unit. Using 2 GNN layers as baseline.}
\label{tab:gpu_mem_cost_2layer}
\end{table*}

\begin{table*}[ht]
\footnotesize
\centering
\begin{tabular}{lccccc}
\toprule
Model & DD & ENZYMES & BZR & COX2 & MUTAG \\
\midrule
FC\textsubscript{cnn}GCN & 209.79$\pm$69.98 & 32.94$\pm$4.76 & 32.64$\pm$0.07 & 32.27$\pm$3.08 & 23.25$\pm$1.17 \\
FC\textsubscript{mlp}GCN & 270.16$\pm$12.61 & 32.10$\pm$5.41 & 28.92$\pm$4.20 & 29.84$\pm$4.41 & 20.80$\pm$1.55 \\
GCN & 166.79$\pm$56.67 & 30.70$\pm$4.35 & 26.41$\pm$3.33 & 31.53$\pm$0.23 & 18.90$\pm$0.82 \\
\midrule
FC\textsubscript{cnn}GIN & 178.01$\pm$37.70 & 35.65$\pm$2.69 & 28.23$\pm$1.41 & 37.58$\pm$0.06 & 26.10$\pm$2.06 \\
FC\textsubscript{mlp}GIN & 103.77$\pm$18.48 & 30.61$\pm$3.55 & 25.05$\pm$4.05 & 33.88$\pm$0.09 & 21.28$\pm$1.83 \\
GIN & 120.80$\pm$46.74 & 31.45$\pm$0.62 & 23.14$\pm$4.48 & 30.70$\pm$4.90 & 23.49$\pm$1.66 \\
\midrule
FC\textsubscript{cnn}GAT & 514.89$\pm$194.63 & 48.83$\pm$10.77 & 40.76$\pm$7.61 & 38.38$\pm$7.55 & 24.86$\pm$1.01 \\
FC\textsubscript{mlp}GAT & 580.53$\pm$126.76 & 50.97$\pm$10.54 & 37.25$\pm$9.35 & 42.37$\pm$10.46 & 24.43$\pm$0.10 \\
GAT & 369.63$\pm$143.28 & 48.51$\pm$13.18 & 24.90$\pm$0.05 & 25.81$\pm$0.03 & 24.30$\pm$4.22 \\
\midrule
FC\textsubscript{cnn}GraphSAGE & 185.61$\pm$35.13 & 32.63$\pm$2.05 & 29.58$\pm$2.66 & 32.70$\pm$1.92 & 23.30$\pm$1.30 \\
FC\textsubscript{mlp}GraphSAGE & 140.55$\pm$39.17 & 28.31$\pm$2.98 & 23.16$\pm$3.85 & 26.13$\pm$4.19 & 21.11$\pm$2.06 \\
GraphSAGE & 118.86$\pm$33.06 & 27.69$\pm$2.84 & 22.89$\pm$3.62 & 26.85$\pm$4.69 & 19.79$\pm$1.66 \\
\midrule
FC\textsubscript{cnn} & 115.57$\pm$9.45 & 25.83$\pm$0.00 & 24.35$\pm$0.00 & 24.89$\pm$0.00 & 21.72$\pm$0.00 \\
FC\textsubscript{mlp} & 31.48$\pm$0.70 & 17.81$\pm$0.20 & 17.97$\pm$0.04 & 18.23$\pm$0.04 & 17.78$\pm$0.00 \\
\bottomrule
\toprule
Model & NCI1 & NCI109 & PROTEINS & PROTEINS\_full & PTC\_MR \\
\midrule
FC\textsubscript{cnn}GCN & 29.33$\pm$3.02 & 28.62$\pm$2.61 & 37.81$\pm$8.05 & 36.87$\pm$8.92 & 22.39$\pm$0.91 \\
FC\textsubscript{mlp}GCN & 30.62$\pm$0.18 & 29.18$\pm$2.83 & 24.39$\pm$0.40 & 31.63$\pm$8.24 & 19.86$\pm$1.99 \\
GCN & 27.41$\pm$3.29 & 27.32$\pm$3.21 & 31.22$\pm$7.21 & 31.26$\pm$7.50 & 20.41$\pm$2.30 \\
\midrule
FC\textsubscript{cnn}GIN & 34.60$\pm$0.11 & 35.75$\pm$0.12 & 34.37$\pm$5.16 & 36.51$\pm$6.64 & 23.56$\pm$1.71 \\
FC\textsubscript{mlp}GIN & 30.36$\pm$3.53 & 32.06$\pm$0.16 & 28.03$\pm$6.60 & 30.41$\pm$5.30 & 20.79$\pm$1.92 \\
GIN & 26.45$\pm$4.16 & 29.67$\pm$3.29 & 30.20$\pm$8.00 & 27.77$\pm$6.46 & 22.04$\pm$2.23 \\
\midrule
FC\textsubscript{cnn}GAT & 38.45$\pm$8.51 & 48.53$\pm$0.47 & 71.10$\pm$21.55 & 74.62$\pm$17.61 & 27.45$\pm$3.84 \\
FC\textsubscript{mlp}GAT & 41.00$\pm$7.50 & 40.92$\pm$7.54 & 62.58$\pm$20.78 & 62.00$\pm$20.95 & 27.44$\pm$4.74 \\
GAT & 43.36$\pm$5.92 & 46.27$\pm$0.61 & 74.43$\pm$19.90 & 68.43$\pm$21.64 & 30.62$\pm$0.32 \\
\midrule
FC\textsubscript{cnn}GraphSAGE & 30.49$\pm$1.82 & 31.01$\pm$2.35 & 30.11$\pm$1.51 & 34.27$\pm$4.69 & 25.04$\pm$0.07 \\
FC\textsubscript{mlp}GraphSAGE & 26.63$\pm$3.80 & 29.80$\pm$0.26 & 25.00$\pm$2.04 & 24.79$\pm$1.87 & 21.70$\pm$1.90 \\
GraphSAGE & 26.19$\pm$3.60 & 22.37$\pm$3.49 & 24.88$\pm$5.65 & 24.95$\pm$5.78 & 20.24$\pm$2.04 \\
\midrule
FC\textsubscript{cnn} & 23.89$\pm$0.48 & 25.06$\pm$0.18 & 26.37$\pm$0.32 & 26.37$\pm$0.32 & 19.88$\pm$0.20 \\
FC\textsubscript{mlp} & 18.00$\pm$0.13 & 18.09$\pm$0.13 & 17.87$\pm$0.06 & 17.94$\pm$0.16 & 17.68$\pm$0.16 \\
\bottomrule
\end{tabular}
\caption{Comparison of model GPU memory usage across 10 datasets (DD, ENZYMES, BZR, COX2, MUTAG, NCI1, NCI109, PROTEINS, PROTEINS\_{full}, PTC\_MR), reported in MB unit. Using 3 GNN layers as baseline.}
\label{tab:gpu_mem_cost_3layer}
\end{table*}

\section{Related Work}

We have not found prior work that directly incorporates functional connectivity into GNN architectures. Accordingly, this section reviews (i) applications of GNNs to brain network analysis using functional connectivity, and (ii) methodological advances in current GNN architectures.

Recent studies have primarily used GNNs to analyze fMRI-derived brain graphs and aid in the diagnosis of neurological disorders, especially in the categorization of disease stages. Several approaches have been proposed, each with unique contributions. Wang et al. introduced IFC-GNN, which integrates multimodal data and functional connectivity to improve Autism Spectrum Disorder (ASD) detection \cite{wang2024ifc}. Kim and Ye leveraged GINs for resting-state fMRI analysis to explore functional connectivity patterns \cite{kim2020understanding}. Klepl et al. utilized EEG-based GNNs for Alzheimer’s disease classification, while Lei et al. applied graph convolutional networks to study functional dysconnectivity in schizophrenia \cite{lei2022graph}. From a methodological perspective, existing approaches can be broadly categorized into two main directions: (1) multimodal data integration and (2) functional connectivity analysis. 

Despite these advancements, existing studies are often limited to specific datasets and tasks, highlighting the need for more generalized approaches that can unify structural and functional connectivity paradigms across broader graph domains. 

Beyond brain network analysis, our work builds upon recent advances in addressing fundamental limitations of GNNs, particularly the over-squashing phenomenon where information from distant nodes becomes compressed through bottleneck structures. Several innovative approaches have emerged to tackle this challenge. DIGL \cite{gasteiger2019digl} employs diffusion-based rewiring that computes a kernel evaluation of the adjacency matrix followed by sparsification, effectively creating shortcuts for information flow. SDRF \cite{topping2021sdrf} takes a geometric approach by identifying edges with low Ricci curvature—indicators of bottleneck locations—and surgically adding edges to support these critical pathways. FoSR \cite{karhadkar2022fosr} formulates the problem as spectral optimization, using iterative first-order methods to maximize the spectral gap and prevent over-squashing. Most recently, PANDA \cite{choi2024panda} introduces an architectural innovation through expanded width-aware message passing, where high-centrality nodes that potentially cause over-squashing are selectively given larger hidden dimensions to accommodate the increased information flow. While these methods successfully address structural limitations through graph rewiring or architectural modifications, our FC-GNN takes an orthogonal approach by augmenting GNNs with explicit topological features inspired by functional connectivity, preserving the original graph structure while capturing complementary global patterns.

\end{document}